%% file: main.tex
\documentclass{article}

\PassOptionsToPackage{numbers,super,sort&compress}{natbib}
\usepackage[preprint]{neurips_2026}

\usepackage[utf8]{inputenc}
\usepackage[T1]{fontenc}
\usepackage{times}
\usepackage{hyperref}
\usepackage{url}
\usepackage{booktabs}
\usepackage{amsfonts}
\usepackage{amsmath}
\usepackage{microtype}
\usepackage{xcolor}
\usepackage{graphicx}
\makeatletter
\@ifundefined{KV@Gin@alt}{\define@key{Gin}{alt}{}}{}
\makeatother
\usepackage{multirow}
\newcommand{\tblcomma}{,}
\usepackage{float}
\usepackage{array}
\usepackage[font=small,labelfont=bf]{caption}
\usepackage{setspace}
\usepackage{fvextra}
\usepackage{newunicodechar}
\newunicodechar{→}{\ensuremath{\rightarrow}}
\newcolumntype{P}[1]{>{\raggedright\arraybackslash}p{#1}}

\hypersetup{colorlinks=true, linkcolor=black, citecolor=black, urlcolor=blue}

\input{tide2_macros.tex}

\input{derived_macros.tex}

\title{TIDE 2.0: an open, model-agnostic engine for keyed
de-identification of clinical notes}

\author{%
  Jose D. Posada\thanks{Corresponding author: \texttt{jdposada@stanford.edu}} \quad
  Somalee Datta \quad
  Priya Desai \\
  \normalfont Technology \& Digital Solutions, Stanford Medicine, Stanford, CA, USA
}
\date{}

\begin{document}

\maketitle

\begin{abstract}
\noindent
Clinical notes capture most of what is documented about a patient's care, but they
cannot be used for research until protected health information (PHI) is removed.
De-identification is often treated as a detection problem. Detection alone is not
sufficient: redaction strips clinical content along with identifiers, date blanking
destroys the temporal intervals needed for longitudinal analysis, and assigning a fresh
random surrogate at each occurrence breaks links between a patient's notes.
We present TIDE 2.0, an MIT-licensed engine with two separable stages: an interchangeable
recognizer and a keyed anonymizer. Both run on hardware the institution owns.
Surrogates are generated cryptographically with no stored linkage table. Dates shift by
a per-patient, interval-preserving offset; each value receives the same surrogate across
all occurrences under a given key; and a release produced under a new key cannot be
linked to earlier releases. We also release \sentryName{}, a recognizer distilled from a
large language model. On two gold-annotated corpora from two institutions, the default
configuration reached span-level recall of \defaultMicroR{} in-domain and
\defaultMicroRItwo{} on the second institution's corpus, at precision \defaultMicroP{}
and \defaultMicroPItwo{}. We report recall and precision per category alongside these
aggregates. The engine is open source, and the recognizer is available under a gated
research-use agreement, so institutions can run, inspect and extend both within their
own environments.
\end{abstract}

\section*{Introduction}
\label{sec:intro}

Much of what is documented about a patient's care is written as free text. Some
diagnoses never reach a billing code. The reasoning behind a treatment change and the
symptoms that precede a diagnosis by years are also often recorded only in the clinical
note. These details are absent from the structured tables held alongside the note in a
research data warehouse
\citep{datta2020newparadigmacceleratingclinical,callahan2023starr}.

Note text supports several secondary uses. A systematic review of case detection found
that reading note text alongside diagnosis codes improves ascertainment
\citep{ford2016extracting}. Postoperative complications appear in operative and progress
notes that discharge coding does not record \citep{murff2011automated}. Off-label
prescribing appears in note text because no structured field encodes indication
\citep{jung2014automated}. Adapted language models have produced summaries of clinical
text that clinicians rated above human-written summaries \citep{vanveen2024adapted}.
Longitudinal questions also depend on the text. Examples are predicting disease from a
patient's trajectory \citep{li2025caread} and reasoning over a clinical history that
spans years \citep{kruse2025temporal}. Both require knowing what happened, when, to whom
and in what order.

Notes cannot be used for research until identifiers are removed. HIPAA Safe Harbor lists
18 categories of protected health information (PHI). These must be handled before a note
can be used for secondary research. A missed identifier can result in a reportable
disclosure. The alternative route, expert determination, requires a statistical
judgement by a qualified expert \citep{hhs2012guidance}. In practice, notes are often
kept behind an access committee or released only as small curated extracts. This limits
the assembly of large corpora.

De-identification has largely been treated as a detection problem, and detection has
improved over time. The first shared task on the problem showed that automatic detection
was feasible on discharge summaries \citep{uzuner2007evaluating}. A review of the
following decade described a field converging on hybrid rule-based and learning systems
\citep{meystre2010automatic}. Sequence-labeling taggers have progressed from conditional
random fields through LSTMs to transformers that generalise across institutions
\citep{chambon_automated_2023}. Recent clinical transformers match or exceed commercial
cloud vendors on span-level accuracy \citep{prakash_improving_2025}. Large language
models perform well without task-specific training, in both zero-shot
\citep{liu_deid-gpt_2023} and few-shot \citep{shield2025} settings.

A detected span is not a de-identified corpus. Redacting every detected span removes
clinical content along with the identifier. In a physician review, most false-positive
detections changed the clinical meaning of a note. Detection F1 did not reflect this
loss \citep{aghakasiri2025overredaction}. Blanking or coarsening dates removes the
intervals that longitudinal analysis relies on. Drawing an independent random surrogate
at each occurrence removes cross-note linkage. A patient, provider or encounter can then
no longer be followed through a record.

Replacing identifiers with realistic surrogates predates most of this detection work.
Scrub replaced each identifier with a realistic value of the same type in 1996
\citep{sweeney1996scrub}. Resynthesised identifiers were later shown not to degrade
taggers trained on the resulting text \citep{yeniterzi2010effects}. Replacement has also
been described as a defence in its own right. A missed identifier is harder to find in a
note containing plausible surrogates than in a note containing redaction marks
\citep{carrell2013hiding}. The i2b2 2014 corpus \citep{stubbs2015corpus} and MC-MED
\citep{kansal2025mcmed} were released with surrogate replacement. MIMIC instead left
bracketed placeholders where identifiers had been \citep{johnson2016mimiciii}. To our
knowledge, few open tools generate surrogates that are both realistic and consistent
across a patient's notes.

The tooling gap also concerns how the parts fit together. The open clinical
de-identifiers were built to process one note at a time. The PhysioNet software is
rule-based and distributed as a script, with no execution layer around it
\citep{neamatullah2008automated}. BoB reached
high recall on Veterans Health Administration documents. It combined rule-based and
statistical components tuned to one institution's document set
\citep{ferrandez2013bob}. Presidio is a widely used PII framework. It ships no clinical
models and no distributed execution layer \citep{presidio2024}. TIDE 1.0, the prior
Stanford tool, couples its recognizer to its anonymizer, so neither can be replaced
independently. It either redacts most structured identifiers or replaces them with
random values \citep{tide2025}. Managed cloud services provide detection but require a
business associate agreement (BAA) with the provider and send clinical text outside the
institution.

To our knowledge, no open engine combines three properties. First, the recognizer can be
exchanged without changing anything else; it may be a large language model (LLM) or a
fine-tuned small language model such as DeBERTa. Second, the engine runs on hardware the
institution already owns. Third, it replaces the detected spans with realistic
surrogates that are consistent across a patient's notes, so that the corpus remains
usable. Supplementary Tables~S7, S10 and S11 compare TIDE 2.0 with TIDE 1.0, OpenMed
and three managed cloud services on these properties.

\begin{figure}[!htb]
  \centering
  \includegraphics[width=0.92\textwidth,alt={Four stacked bands showing one clinical sentence transformed down the page.
  Band 1, the note as written, names a patient, a medical record number, a physician,
  two dates, a hospital and a phone number. Band 2 shows the same sentence with each identifier
  bracketed and typed as PATIENT, ID, DOCTOR, DATE, HOSPITAL or PHONE.
  Band 3 lists the replacement strategy per category: keyed selection from a surrogate list for names,
  FF3 format-preserving encryption for identifiers and phone numbers, and a per-patient HMAC date
  offset that preserves the fourteen-day interval. Band 4 shows the released note, in which every
  identifier has been replaced by a plausible surrogate of the same type and both dates have moved by the same ten days.}]
  {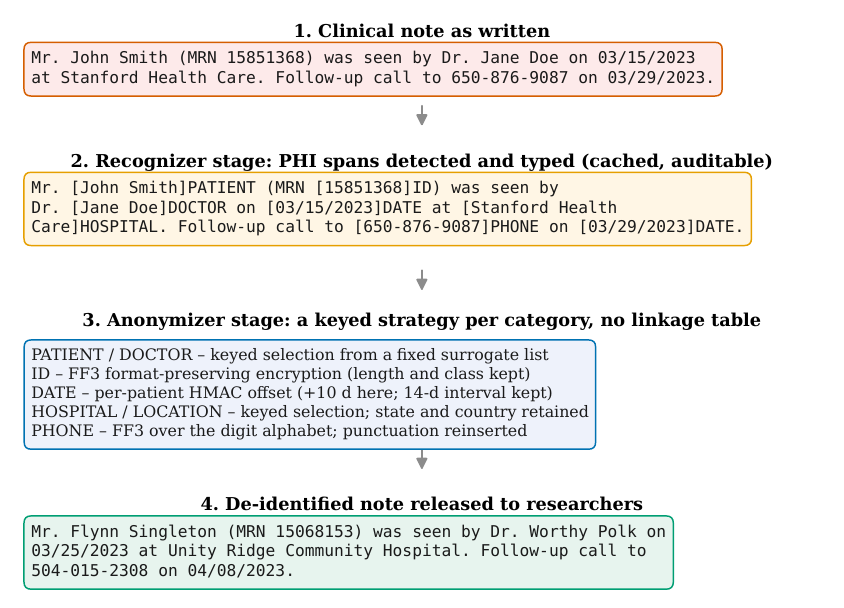}
  \caption{\textbf{One sentence through the pipeline.} The recognizer stage detects and
  types PHI spans and caches them before any text is altered. A reviewer can therefore
  inspect the detected spans before the anonymizer runs. The anonymizer stage applies a
  keyed strategy to each category. Both dates move by the same per-patient offset, so the
  fourteen-day interval between them is preserved. Fig.~\ref{fig:architecture} shows the
  component-level architecture.}
  \label{fig:pipeline}
\end{figure}

We built this engine. TIDE 2.0 \footnote{\tidetworepo} is an MIT-licensed framework that
separates a recognizer stage from an anonymizer stage (Fig.~\ref{fig:pipeline}). A site
can therefore adopt a recognizer today and replace it later. Notes that were already
processed do not need to be re-anonymized. Both stages run on Ray, a distributed
runtime, and can read from and write to warehouse note tables. In this paper we
benchmark them on a single commodity GPU; we do not report a warehouse-scale deployment.

The anonymizer is keyed and table-free. Dates move by a per-patient offset that
preserves every interval. Alphanumeric identifiers are replaced by format-preserving
encryption, and names are replaced by deterministic keyed selection. The same value
therefore receives the same surrogate everywhere under one key, and no database of
mappings is stored. A release produced under a new key cannot be linked to releases
produced under the old one.

The default configuration pairs a token classifier, distilled from an LLM, with a regex
layer, and runs behind the institutional firewall. Any other model can fill the
recognizer stage. We report what the default configuration detects on two gold-annotated
corpora from two institutions and
where, per category, it fails. We also report what the anonymizer guarantees and its
throughput on one GPU. We release the engine and the recognizer so that other sites can
run, inspect and extend them.

\section*{Results}
\label{sec:results}

\subsection*{Study design}

We evaluated every recognizer on two gold-annotated corpora. SHIELD
\citep{shield2025} contains \shieldNotes{} clinical notes from Stanford Medicine with
\shieldGoldSpans{} annotated PHI spans in \shieldNumCats{} categories, a mean of
\shieldSpansPerNote{} identifiers per note. The i2b2 2014 de-identification corpus
\citep{stubbs2015corpus} contains \itwoNotes{} clinical notes from a different
institution. After the label crosswalk described in Methods, it carries \itwoGoldSpans{}
spans, a mean of \itwoSpansPerNote{} identifiers per note. The two corpora differ in
institution, note type and identifier density. SHIELD measures in-domain behaviour, and
i2b2 measures what a site inherits when it adopts a model trained elsewhere.

The primary outcome is span-level recall, reported per category and per corpus. We treat
a missed identifier as a
disclosure, which differs from a precision error that removes a clinical word. Secondary
outcomes are span-level precision, sustained throughput and the structural guarantees of
the anonymizer. Precision indicates how much clinical content survives replacement.
Methods gives the span-matching rule, the \overlapThresh{} overlap threshold and the
bootstrap procedure.

Table~\ref{tab:detection} reports span-level precision and recall on both corpora. It
covers the default configuration (\sentryName{}\texttt{+regex}), \sentryName{} alone,
regex alone, TIDE 1.0, the two AIMI recognizers and the \numExternalPII{} external
open-source PII taggers.

\begin{table}[!htb]
  \caption{\textbf{PHI detection performance by recognizer and corpus.} Span-level
  precision (P) and recall (R) are micro-averaged over pooled category counts and
  macro-averaged over the \shieldNumCats{} categories, at an overlap threshold of
  \overlapThresh{}. The upper block shows the regular-expression and gazetteer layer
  alone, TIDE 1.0, the two AIMI recognizers, and the distilled \sentryName{}
  with and without the regex layer; \sentryName{}\texttt{+regex} is the default
  configuration. The lower block shows the \numExternalPII{} external
  open-source PII taggers. TIDE 1.0 was run on SHIELD only (``--''). The AIMI
  recognizers have no AGE, LOCATION or WEB label, and their macro averages include
  zeros for these three categories. The AIMI recognizers were also trained on i2b2
  2014, so that corpus is not held out for them. No other model was trained on either
  corpus.
  Per-category values
  are in Supplementary Tables~S1--S2. The same values with bootstrap confidence
  intervals are in Supplementary Tables~S4--S5.}
  \label{tab:detection}
  \centering
  \small
  \setlength{\tabcolsep}{4.5pt}
  \begin{tabular}{lcccccccc}
    \toprule
    & \multicolumn{4}{c}{\textbf{SHIELD}} & \multicolumn{4}{c}{\textbf{i2b2 2014}} \\
    \cmidrule(lr){2-5} \cmidrule(lr){6-9}
    & \multicolumn{2}{c}{micro} & \multicolumn{2}{c}{macro}
    & \multicolumn{2}{c}{micro} & \multicolumn{2}{c}{macro} \\
    \cmidrule(lr){2-3} \cmidrule(lr){4-5} \cmidrule(lr){6-7} \cmidrule(lr){8-9}
    \textbf{Recognizer} & P & R & P & R & P & R & P & R \\
    \midrule
    TIDE2 regex & 0.97 & 0.42 & 0.52 & 0.39 & 0.98 & 0.39 & 0.48 & 0.27 \\
    tide1 & 0.48 & 0.39 & 0.41 & 0.38 & -- & -- & -- & -- \\
    AIMI v1 & 0.67 & 0.72 & 0.45 & 0.55 & 0.73 & 0.67 & 0.44 & 0.49 \\
    AIMI v2 & 0.63 & 0.70 & 0.43 & 0.53 & 0.70 & 0.65 & 0.43 & 0.47 \\
    TIDE2-Sentry & 0.89 & 0.88 & 0.85 & 0.81 & 0.87 & 0.77 & 0.72 & 0.67 \\
    TIDE2-Sentry + regex & 0.88 & 0.88 & 0.83 & 0.80 & 0.87 & 0.77 & 0.72 & 0.67 \\
    \midrule
    \multicolumn{9}{l}{\emph{External open-source PII taggers}} \\
    DeBERTa-PII (lakshyakh93) & 0.24 & 0.30 & 0.19 & 0.35 & 0.14 & 0.18 & 0.10 & 0.21 \\
    OpenMed SuperClinical L & 0.49 & 0.48 & 0.48 & 0.47 & 0.54 & 0.45 & 0.43 & 0.41 \\
    OpenMed SuperClinical S & 0.50 & 0.51 & 0.49 & 0.54 & 0.53 & 0.46 & 0.40 & 0.43 \\
    PII-Identifier (antonio) & 0.50 & 0.48 & 0.50 & 0.49 & 0.57 & 0.42 & 0.51 & 0.43 \\
    Ettin-68M Nemotron & 0.52 & 0.48 & 0.53 & 0.50 & 0.57 & 0.45 & 0.44 & 0.44 \\
    ekacare MiniLM & 0.43 & 0.46 & 0.45 & 0.58 & 0.50 & 0.48 & 0.33 & 0.46 \\
    \bottomrule
  \end{tabular}
\end{table}

On SHIELD, the default configuration reaches micro-averaged recall \defaultMicroR{} at
precision \defaultMicroP{}; macro-averaged recall and precision are \defaultMacroR{} and
\defaultMacroP{}. \sentryName{} alone reaches micro-averaged recall \sentryMicroR{} at
precision \sentryMicroP{}, and macro-averaged recall and precision of \sentryMacroR{} and
\sentryMacroP{}. The stronger of the two AIMI recognizers \citep{prakash_improving_2025},
\bestOtherName{}, reaches micro-averaged recall \bestOtherMicroR{} at precision
\bestOtherMicroP{}. The AIMI recognizers were trained on radiology reports and on
i2b2 2014. Their label set covers \aimiCats{} and has no AGE, LOCATION or WEB label, so
they score zero in those categories. Over the six categories they cover, macro-averaged
recall is \aimiOneMacroRAimiCats{} for \bestOtherName{} and \sentryMacroRAimiCats{} for
\sentryName{}. Both AIMI recognizers match or exceed \sentryName{} on DATE and ID.
TIDE 1.0 reaches micro-averaged recall \tideOneMicroR{} at precision \tideOneMicroP{}.

On i2b2 2014, the default configuration reaches micro-averaged recall
\defaultMicroRItwo{} at precision \defaultMicroPItwo{}, and \sentryName{} alone reaches
\sentryMicroRItwo{} at \sentryMicroPItwo{}. Macro-averaged recall and precision of
\sentryName{} are \sentryMacroRItwo{} and \sentryMacroPItwo{}, or \sentryMacroRSupItwo{}
for recall when WEB, with \itwoWebSupportItwo{} gold spans, is left out.
\bestOtherNameItwo{} reaches micro-averaged
recall \bestOtherMicroRItwo{} at precision \bestOtherMicroPItwo{}. This corpus is
in-domain for the AIMI recognizers and out-of-domain for \sentryName{}.
Over the six categories in the AIMI label set, macro-averaged recall is
\aimiOneMacroRAimiCatsItwo{} for \bestOtherNameItwo{} and \sentryMacroRAimiCatsItwo{} for
\sentryName{}. \bestOtherNameItwo{}
has the higher recall in \bestOtherCatsLedItwo{} of the \shieldNumCats{} categories.
The advantage of \sentryName{} is concentrated in the categories the AIMI recognizers
do not cover (AGE, LOCATION) or detect poorly (DOCTOR, PATIENT). On PATIENT,
\sentryName{} reaches \sentryPatientRItwo{} against \bestOtherPatientRItwo{}.

\begin{figure}[!htb]
  \centering
  \includegraphics[width=\textwidth,height=0.78\textheight,keepaspectratio,alt={Two radar plots of span-level recall across nine PHI categories, one for SHIELD and one for i2b2 2014. Each plot shows the regex layer, the two AIMI recognizers, TIDE2-Sentry and six external open-source PII taggers.}]{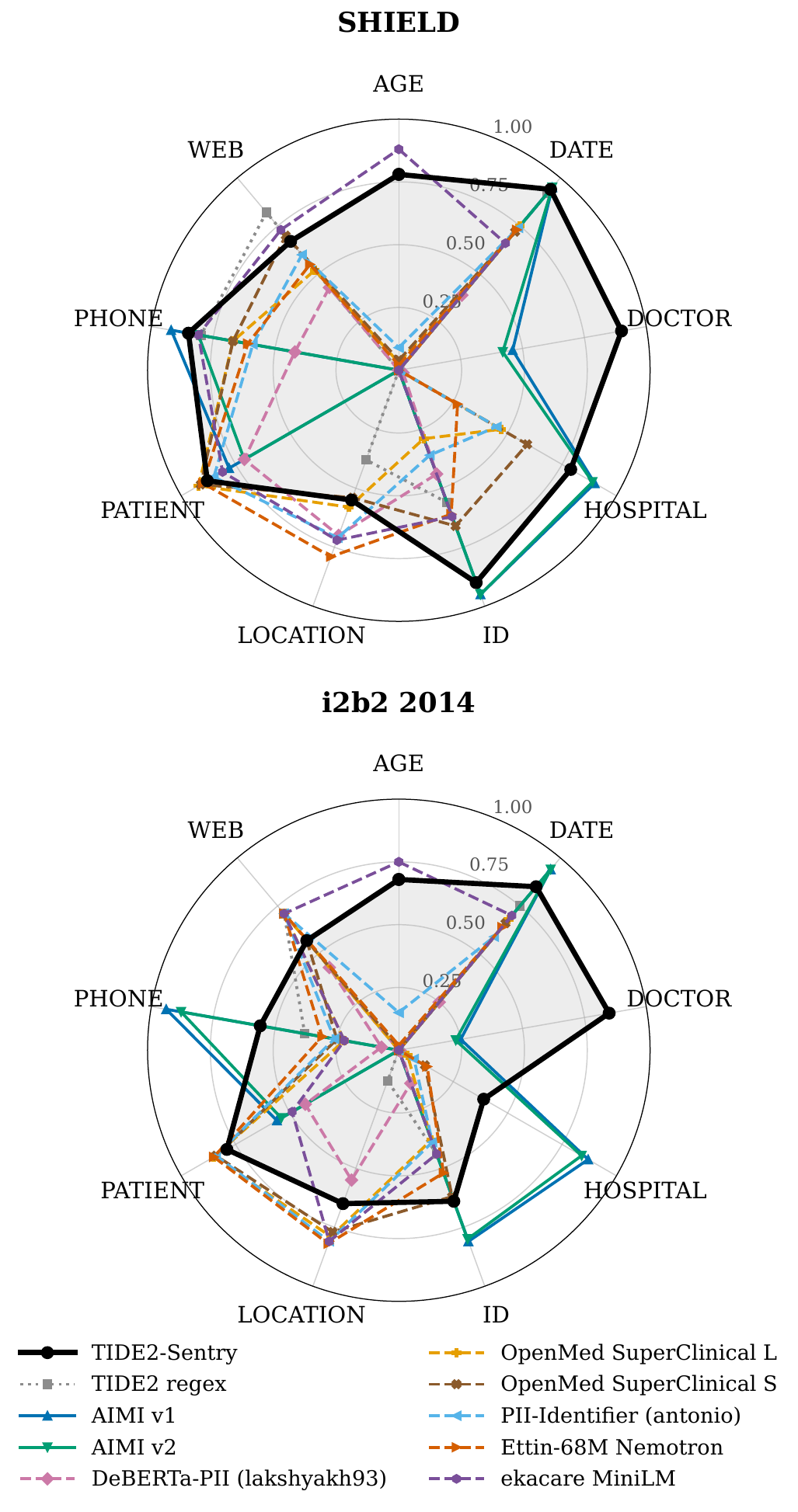}
  \caption{\textbf{Per-category recall of all recognizers on each corpus.} Each axis is
  one PHI category, and the radial scale is span-level recall from 0 to 1 at an overlap
  threshold of \overlapThresh{}. Top: SHIELD. Bottom: i2b2 2014. Each panel shows the
  regex layer, the two AIMI recognizers, \sentryName{} and the \numExternalPII{} external
  open-source taggers. The \texttt{+regex} ablations are omitted. Values are those in
  Supplementary Tables~S1--S2.}
  \label{fig:radar}
\end{figure}

The \numExternalPII{} external open-source PII taggers were run on both corpora. They
were trained on non-clinical or non-US identifier distributions, so they serve as an
out-of-distribution probe and not as tuned baselines. Their micro-averaged recall ranges
from \extMicroRMin{} to \extMicroRMax{} on SHIELD and from \extMicroRMinItwo{} to
\extMicroRMaxItwo{} on i2b2 2014. The highest values are \extBestName{} on SHIELD
(precision \extMicroPBest{}) and \extBestNameItwo{} on i2b2 (precision
\extMicroPBestItwo{}). The maximum DOCTOR recall among the external taggers is
\extDoctorRMax{} on SHIELD and \extDoctorRMaxItwo{} on i2b2. Their taxonomies have a
single generic person type and do not separate patients from providers, which clinical
de-identification requires. When DOCTOR and PATIENT are pooled into one NAME category,
their NAME recall ranges from \extNameRMin{} to \extNameRMax{} on SHIELD and from
\extNameRMinItwo{} to \extNameRMaxItwo{} on i2b2, against \sentryNameR{} and
\sentryNameRItwo{} for \sentryName{} (Supplementary Table~S14). Figure~\ref{fig:radar}
shows per-category recall for all recognizers on both corpora.

Regex alone reaches ID recall \regexIdR{}, \sentryName{} alone reaches \sentryIdR{}, and
the two together reach \sentryPlusRegexIdR{}. Adding the regex layer changes ID recall
by \regexGainIdR{} (95\% CI \regexGainIdRCI{}) on SHIELD and \regexGainIdRItwo{}
(\regexGainIdRCIItwo{}) on i2b2, and micro-averaged recall by \regexGainMicroR{}
(\regexGainMicroRCI{}) and \regexGainMicroRItwo{} (\regexGainMicroRCIItwo{}). It also
lowers LOCATION recall and WEB precision on SHIELD (Supplementary Table~S13). On these
two corpora, the net effect of the regex layer on detection is therefore small.

Moving from SHIELD to i2b2 2014 reduces the micro-averaged recall of \sentryName{} by
\sentryMicroRDropItwo{}, from \sentryMicroR{} to \sentryMicroRItwo{}. Precision changes
little, from \sentryMicroP{} to \sentryMicroPItwo{}.

Supplementary Table~S12 splits the drop into one term per category. Each term weights a
category's recall change by its share of gold spans (Methods). Recall changes account for
\decompRateTotal{} of the drop, and the different category mix of the two corpora for
\decompMixTotal{}. HOSPITAL, ID and PHONE together contribute \decompVocabShare{} of the
drop (HOSPITAL \decompHosp{}, ID \decompId{}, PHONE \decompPhone{}). HOSPITAL recall falls
by \sentryHospRDropItwo{}, from \sentryHospR{} to \sentryHospRItwo{}; PHONE by
\sentryPhoneRDropItwo{}, from \sentryPhoneR{} to \sentryPhoneRItwo{}; and ID by
\sentryIdRDropItwo{}, from \sentryIdR{} to \sentryIdRItwo{}. DATE contributes
\decompDateShare{} (\decompDate{}). Its recall falls less, from \sentryDateR{} to
\sentryDateRItwo{}, but DATE is the largest category on i2b2. DOCTOR goes from
\sentryDoctorR{} to \sentryDoctorRItwo{}. LOCATION recall is higher on i2b2 than on
SHIELD, \sentryLocRItwo{} against \sentryLocR{}. WEB has \itwoWebSupportItwo{} gold spans
on i2b2, and we do not interpret its change.

We did not analyse the missed spans, so the causes of these drops are hypotheses. One
hypothesis is that HOSPITAL, ID and PHONE depend on local vocabulary and formats, such
as facility names and acronyms, identifier formats and phone conventions, that are rare
in the Stanford training notes. The drop also mixes other differences between the
corpora: note types, the surrogate identifiers that replace real ones in i2b2 (Methods),
and the label crosswalk.

We also varied the scoring choices. Re-mapping isolated years,
weekdays, seasons and bare month names in i2b2 from PHI to safe, following the
convention of Philter, changes the i2b2 recall and precision of \sentryName{} by less
than 0.03, in opposite directions. Trimming leading and trailing whitespace from
predicted spans before matching raises i2b2 recall by 0.03 and SHIELD recall by 0.01.
The crosswalk decisions were made with access to both corpora. Without the boundary
alignment of the crosswalk, i2b2 micro-averaged recall of the default configuration is
\defaultMicroRNoAlignItwo{} rather than \defaultMicroRItwo{}. At an overlap threshold of
0.5 instead of \overlapThresh{}, it is \defaultMicroRIouHalf{} on SHIELD and
\defaultMicroRIouHalfItwo{} on i2b2. Counted by characters instead of spans and
ignoring category, predicted spans cover \defaultCharR{} of gold PHI characters on
SHIELD and \defaultCharRItwo{} on i2b2. This is the share of identifier text that the
anonymizer rewrites. The Philter re-mapping and whitespace trimming do not change the
ordering of the recognizers or the cross-corpus pattern; the other three checks were run
for the default configuration only. Supplementary Table~S6 records every crosswalk
decision.

\subsection*{Distillation, anonymizer and throughput}

\sentryName{} is a \sentryBase{} token classifier distilled from \teacherName{} over
\teacherLabelNotes{} unannotated notes (Methods). To measure distillation fidelity, the
teacher was scored on all \shieldNotes{} SHIELD notes with the extraction prompt in
Supplementary Note~1. That prompt was developed on a few dozen notes outside SHIELD and
i2b2. Supplementary Fig.~S1 compares teacher and student per category on SHIELD. The
full table with confidence intervals is Supplementary Table~S3.

The student reaches macro-averaged recall \sentryMacroR{}, against \flashMacroR{} for the
teacher. It has lower recall than the teacher in every category. Ordered by size, the
recall differences are WEB (\distWebDelta{}), AGE (\distAgeDelta{}), LOCATION
(\distLocDelta{}), PHONE (\distPhoneDelta{}), HOSPITAL (\distHospDelta{}), PATIENT
(\distPatientDelta{}), DOCTOR (\distDoctorDelta{}), ID (\distIdDelta{}) and DATE
(\distDateDelta{}). After Bonferroni correction across the \shieldNumCats{} categories,
every difference except WEB (adjusted $p\distWebPadj{}$) remains significant at 0.05;
the ID difference is close to that threshold (adjusted $p\distIdPadj{}$). Both models
have their lowest recall on LOCATION: \llmFlashLocR{} for the teacher and \sentryLocR{}
for the student. The low recall in this category is therefore shared with the teacher,
and distillation adds a further gap ($\Delta$R~\distLocDelta{}).

The guarantees of the anonymizer follow from a keyed, table-free construction. The
measurement question is whether the implementation realizes the construction.
Table~\ref{tab:anon-metrics} reports a seeded synthetic stress set of \anonStressN{}
distinct values per keyed category. The set contains no PHI.

\begin{table}[!htb]
  \caption{\textbf{Anonymizer verification.} A seeded synthetic stress set containing no
  PHI (\anonStressN{} distinct values per keyed category). The first four rows are structural guarantees
  of the keyed construction, and they check that the implementation realizes them. The
  remapping row uses 1{,}000 two-token names and 1{,}000 eight-digit identifiers. The
  collision rows measure a property that is not guaranteed a priori: whether
  distinct real values receive distinct surrogates. The 3--5-digit row tests every value
  of each length. $N$ is the number of distinct values tested. These are unit-level
  checks of the implementation on synthetic values, not a re-identification or linkage
  attack.}
  \label{tab:anon-metrics}
  \centering
  \small
  \begin{tabular}{llrl}
    \toprule
    \textbf{Property} & \textbf{Construction} & \textbf{$N$} & \textbf{Result} \\
    \midrule
    Determinism (same key) & keyed HMAC /\allowbreak{} FF3 /\allowbreak{} selection & 2\tblcomma{}000 & 100.0\% identical \\
    Remapping (rotated key) & new key remaps surrogate & 2\tblcomma{}000 & 100.0\% remapped \\
    Cross-note consistency & same value $\to$ same surrogate & 1\tblcomma{}000 & 100.0\% linked \\
    Interval preservation & per-patient date shift & 2\tblcomma{}000 & 100.0\% exact day-delta \\
    ID collisions\tblcomma{} 8 digits & FF3\tblcomma{} no padding & 2\tblcomma{}000 & 0.00\% (2\tblcomma{}000 distinct) \\
    ID collisions\tblcomma{} 3--5 digits & FF3\tblcomma{} padded\tblcomma{} truncated & 111\tblcomma{}000 & 34.8\%--36.6\% \\
    Name collisions & keyed per-token selection; pool 2\tblcomma{}276 & 2\tblcomma{}000 & 0.4\% (1\tblcomma{}993 distinct) \\
    Name collisions & keyed per-token selection; pool 2\tblcomma{}276 & 20\tblcomma{}000 & 3.3\% \\
    \bottomrule
  \end{tabular}
\end{table}

Determinism, cross-note consistency, remapping under a new key and interval
preservation each hold on \anonDetRate{} of trials. This is expected for a keyed
deterministic construction. Supplementary Table~S8 shows worked examples per category,
and Supplementary Table~S9 follows one name, identifier and date across two notes under
a fixed and a new key. The collision measurements are not guaranteed a priori.
Eight-digit identifiers need no padding, and FF3 over a fixed content length is a
permutation; they collided at \anonMrnColl{}. Numeric identifiers shorter than six
digits are padded, encrypted and truncated back to their original length (Methods), so
the map is not a permutation. Across all 3- to 5-digit values, \anonShortIdCollMin{} to
\anonShortIdCollMax{} of values shared a surrogate with another value. Name surrogates
are drawn by keyed selection from a pool of \anonNamePool{} entries per token, so
distinct real names can collide. We measured \anonNameColl{} among \anonStressN{} names
and \anonNameCollLarge{} among \anonNameLargeN{}; the rate grows with the number of
distinct names. A collision merges two identities under one surrogate. This reduces
linkage precision and does not reveal either original value.

Table~\ref{tab:throughput} reports throughput per stage on \thruNumGpu{} \thruGpu{} GPU
with \thruVcpus{} vCPUs, for the full SHIELD corpus and for a bounded subset of the
longest notes. The stages ran as one automatically chained job; we did not measure
whether they overlapped in time. GPU recognition was the slowest stage on both corpora.
Peak GPU memory on the long-note subset was \thruLongVram{}~GB. No batch failed for lack
of memory in either configuration. The CPU-only path was not tuned (Supplementary
Note~3).

\begin{table}[!htb]
  \caption{\textbf{Throughput per stage, in content-tokens per second.}
  \thruNumGpu{}~\thruGpu{} GPU on a \thruPlatform{} with \thruVcpus{}~vCPUs and
  \thruRamGb{}~GB of RAM, running \sentryName{} with a \windowBudget{}-token window
  budget and \windowOverlap{}-token overlap. Transformer: the token classifier.
  Recognizer: the CPU regex and Aho--Corasick stage. Left: the full SHIELD corpus
  (\shieldNotes{} notes, \thruShieldTokens{} tokens). Right: a bounded subset of the
  longest notes (\thruLongTokens{} tokens). One run per configuration, so these are
  indicative rather than a mean. The CPU-only columns are an untuned single-actor
  configuration; the CPU stages are shared between the two paths.}
  \label{tab:throughput}
  \centering
  \small
  \begin{tabular}{lrr@{\hskip 2em}rr}
    \toprule
    & \multicolumn{2}{c}{\textbf{SHIELD}} & \multicolumn{2}{c}{\textbf{Longest notes}} \\
    \cmidrule(lr){2-3} \cmidrule(lr){4-5}
    \textbf{Stage} & \textbf{L4 path} & \textbf{CPU-only} & \textbf{L4 path} & \textbf{CPU-only} \\
    \midrule
    Transformer & 4\tblcomma{}063 & 200 & 10\tblcomma{}104 & 358 \\
    Recognizer & 6\tblcomma{}127 & 6\tblcomma{}127 & 27\tblcomma{}591 & 27\tblcomma{}591 \\
    Anonymizer & 4\tblcomma{}421 & 4\tblcomma{}421 & 21\tblcomma{}547 & 21\tblcomma{}547 \\
    End-to-end & 1\tblcomma{}573 & 186 & 5\tblcomma{}506 & 348 \\
    \bottomrule
  \end{tabular}
\end{table}

\section*{Discussion}
\label{sec:discussion}

The default configuration, a distilled token classifier with a regex layer running behind
an institutional firewall, detected \defaultMicroR{} of the PHI spans in a Stanford corpus
and \defaultMicroRItwo{} of those in a corpus from a different institution. Precision was
\defaultMicroP{} and \defaultMicroPItwo{}, respectively. At the second institution,
HOSPITAL, ID and PHONE contributed \decompVocabShare{} of the loss in recall and DATE
\decompDateShare{}. The anonymizer that consumes the detected spans was deterministic,
interval-preserving and remapped under a new key on \anonDetRate{} of trials in a
synthetic stress set. The full SHIELD corpus ran end to end at \thruShieldTotalLfour{}
content-tokens per second on one commodity GPU. Together, these results show that
detection with an exchangeable recognizer, keyed surrogate replacement, interval-preserving
date shifts and cross-note consistency can be combined on hardware a site owns.

Most de-identification studies report F1, which treats a leaked identifier and a removed
clinical term as equivalent errors. The two errors have different consequences. A missed
identifier is a reportable disclosure, and documented re-identification attacks on health
data mostly start from identifiers that were missed \citep{elemam2011systematic}. A false positive
replaces a clinical word with a surrogate and reduces research signal. One study found
that encoders pretrained on automatically de-identified clinical text lose little on
downstream tasks \citep{vakili2022downstream}. A privacy officer needs to know how many
identifiers remain, and a researcher needs to know how much of each note is preserved. A
single averaged score answers neither question.

F1 also does not measure what happens after detection. A system that detects every
identifier and then deletes it receives a perfect score, yet the resulting corpus
contains no date intervals, patient threads or provider attributions. Redaction-based
tooling behaves this way by default. A physician review found that most false-positive
detections change the clinical meaning of a note, and detection metrics did not register
this \citep{aghakasiri2025overredaction}. We therefore used span-level recall as the
primary outcome, because a missed identifier carries the greater cost. We report
precision as an indicator of the clinical content that survives replacement. We suggest
reporting recall and precision as a pair rather than as their harmonic mean.

Clinical de-identification has a long history. Philter combined regular expressions, a
part-of-speech tagger and safe and unsafe word lists, and it remains a reference point
for recall-first design \citep{norgeot2020philter}. The MIST toolkit treated surrogate
replacement as a primary output \citep{aberdeen2010mist}. Recurrent taggers with a
conditional random field layer, neural sequence labelling and transformers pretrained on
clinical text later improved results on the same benchmark
\citep{liu2017deidentification,dernoncourt2017neuroner,johnson2020deidentification}.
Annotated corpora and shared tasks shaped this progression. The 2014 i2b2/UTHealth task
provided the benchmark and evaluation script \citep{stubbs2015overview}, and the 2016
CEGS N-GRID task moved evaluation to psychiatric intake records
\citep{stubbs2017cegs}. An independently annotated paediatric corpus showed that systems
score below their published results on corpora they were not tuned on
\citep{deleger2013preparing}. More recently, transformers have matched or exceeded
commercial cloud vendors on span-level accuracy
\citep{chambon_automated_2023,prakash_improving_2025}, and large language models perform
well without task-specific training \citep{liu_deid-gpt_2023}.

Our comparison covers the two AIMI recognizers and six external open-source taggers. On
SHIELD, \sentryName{} reached micro-averaged recall \sentryMicroR{}, against
\bestOtherMicroR{} for \bestOtherName{}. On i2b2 2014 the values were
\sentryMicroRItwo{} and \bestOtherMicroRItwo{}. The AIMI recognizers were also trained on
i2b2 2014, so this corpus is in-domain for them and out-of-domain for \sentryName{}.
The i2b2 comparison is therefore not a held-out comparison for AIMI, and we do not use it
to rank the two. \bestOtherNameItwo{} had higher recall than \sentryName{} in
\bestOtherCatsLedItwo{} categories on i2b2: DATE, HOSPITAL, ID and PHONE. One explanation
is that these categories depend on local formats and vocabulary, which a model trained
on the corpus can learn; the same categories account for most of the cross-corpus loss
of \sentryName{}, described below. The external taggers reached micro-averaged recall of
\extMicroRMin{} to \extMicroRMax{} on SHIELD and \extMicroRMinItwo{} to
\extMicroRMaxItwo{} on i2b2. They were not built for clinical notes and serve as a
robustness probe.

We did not compare TIDE 2.0 against Philter, MIST, NeuroNER or the top i2b2 2014
systems. Each of them is a recognizer, and TIDE 2.0 provides the stage into which a
recognizer is placed. Any of them could fill that stage. A comparison among them would
indicate which recognizer a site should choose. It would say nothing about the engine,
the anonymizer or the execution layer, which are the subject of this paper. Because the
recognizer stage is open, such a comparison on one corpus and one metric is
straightforward to set up, and we regard it as a useful next study.

The tools a site might compare with TIDE 2.0 are different in kind. Presidio is the most
widely used open PII framework, and TIDE 2.0 adopts its recognizer interface. It ships no
clinical models and no distributed execution layer, so a site building on it must write
the warehouse tier itself \citep{presidio2024}. OpenMed is a per-document detection
toolkit, and its models can fill the recognizer stage. Managed cloud services detect and,
in most cases, replace identifiers. They require clinical text to leave the institutional
perimeter under a BAA.
TIDE 1.0, the predecessor of this work, coupled recognition to a fixed rule-and-dictionary
engine and redacted most structured identifiers instead of replacing them
\citep{tide2025}. Supplementary Tables~S7, S10 and S11 compare these options with
TIDE 2.0 feature by feature. To our knowledge, none of them combines an exchangeable
recognizer with keyed, cross-note-consistent replacement in an open engine that runs
inside the institution.

TIDE 2.0 is designed to read from and write back to warehouse note tables, but this paper
does not report a warehouse-scale deployment. Its contribution is the separation of an
exchangeable recognizer from a keyed anonymizer, and the release of both the engine and a
distilled recognizer.

\begin{figure}[!htb]
  \centering
  \includegraphics[width=0.92\textwidth,alt={A dashed box labelled institutional security perimeter contains four components: clinical notes from warehouse source tables, a Ray cluster of GPU recognizer actors and CPU anonymizer actors, a cache of PHI spans inspected before any text is altered, and below them key custody holding per-study rotatable salts and keys, and an audit trail of the span cache and configuration. A single arrow leaves the perimeter to a de-identified corpus released to researchers under agreed terms, labelled as the only output that crosses.}]{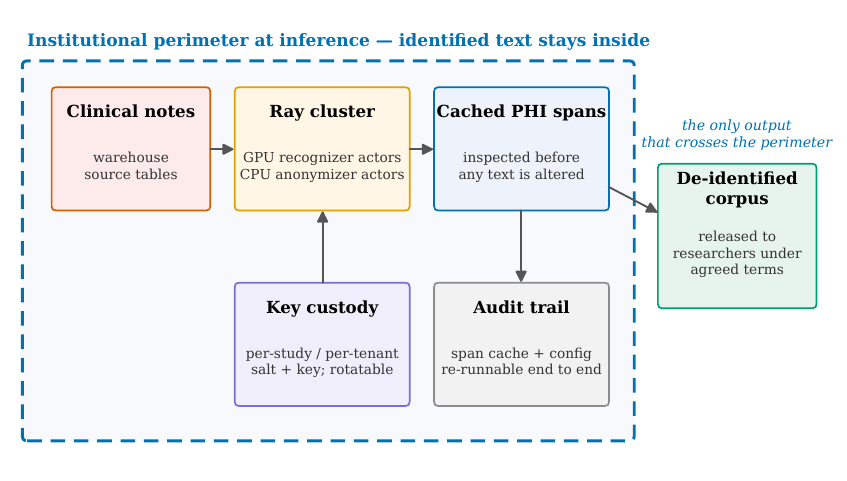}
  \caption{\textbf{Deployment and the institutional perimeter at inference.} All
  components that handle identified text run inside the institution. Keys are held per
  study or tenant. A new key re-maps every surrogate deterministically, so releases
  under different keys cannot be linked; releases already issued are not changed. The
  de-identified corpus is the only artefact that crosses the perimeter.}
  \label{fig:deployment}
\end{figure}

Several governance properties follow from the cryptographic construction. No linkage
table is stored, so there is no table to protect, audit or lose. Consistency across notes
depends on a key. Keys are isolated per study or tenant, so two releases produced under
different keys cannot be linked to each other. A new key re-maps every surrogate
deterministically. Rotation does not protect releases already issued: a party holding a
disclosed key can recompute the surrogates in those releases for any candidate value,
and those releases would need to be withdrawn or re-issued. Detected spans are cached
before any text is altered, so a reviewer can inspect what a model found before the
anonymizer runs. At inference, the released recognizer runs on institutional hardware,
so no protected health information crosses the perimeter, and the de-identified corpus is
the only artefact that leaves it (Fig.~\ref{fig:deployment}). Distillation was the
exception: the teacher labelled identified notes through a cloud service under a BAA
(Methods). The engine is MIT-licensed, and a security reviewer can read its code to
check each of these properties.

The error pattern is more informative than the aggregate metrics. At the second
institution, \sentryName{} lost \sentryHospRDropItwo{} of its HOSPITAL recall,
\sentryPhoneRDropItwo{} of PHONE and \sentryIdRDropItwo{} of ID. Weighted by their share of
gold spans, these three categories contributed \decompVocabShare{} of the drop in
micro-averaged recall, and DATE, the largest category on i2b2, contributed
\decompDateShare{}. LOCATION improved. We hypothesise that the three categories with
large drops depend on local vocabulary, such as facility acronyms, phone conventions and
identifier formats, but we did not analyse the missed spans to confirm this.

LOCATION is a different case, and we have not yet corrected it. It is the weakest
category for the student (\sentryLocR{} on SHIELD) and for the teacher
(\llmFlashLocR{}). The low recall is shared with the teacher, which suggests that the
label definition and the surface diversity of the category contribute; distillation adds
a further gap ($\Delta$R~\distLocDelta{}). The two AIMI recognizers have no AGE, LOCATION
or WEB label, so they cannot detect these categories on either corpus. This is a
coverage limit of their label set, not a domain shift. The external PII taggers scored
near zero on DOCTOR on both corpora. Their taxonomies have one generic person type and do
not separate patients from providers; with DOCTOR and PATIENT pooled, their NAME recall
reaches up to \extNameRMax{} on SHIELD. These three results have different causes, which
supports reading per-category recall rather than an aggregate.

This study has limitations. We did not run a downstream phenotyping or temporal-reasoning
experiment on the de-identified corpus. Utility preservation is therefore argued by
construction and verified structurally, and downstream equivalence is not demonstrated.
Two designs could address this: a direct retention measurement in the style of the CIRE
metric \citep{aghakasiri2025overredaction}, and the pretrain-and-evaluate protocol used
to show that automatic de-identification leaves downstream accuracy largely intact
\citep{vakili2022downstream}.

Throughput comes from single runs on one machine and is indicative rather than a mean.
The CPU-only path is an untuned single-actor configuration. We did not run on more than
one node, measure overlap between stages, or test recovery from a failed worker.

The student was trained on \teacherLabelNotes{} teacher-generated silver labels. We did
not characterise the noise in these labels against gold, so the effect of the teacher's
systematic errors on the student is unquantified. The distillation notes are disjoint
from both corpora at the note level; we did not verify that their patients are disjoint
from SHIELD's. The AIMI recognizers were trained on i2b2 2014, so i2b2 is not a held-out
corpus for them, and we did not verify that their Stanford training data are disjoint
from SHIELD. The identifiers in i2b2 are surrogates, so recall there may differ from
recall on real identifiers from that institution. The crosswalk decisions were made with
access to both corpora. The \numExternalPII{} external
taggers are off-distribution on clinical notes and serve as a robustness probe, not as
tuned baselines. We do not report an LLM recognizer comparison. Prompt development was
carried out for the teacher model only, and a comparison with unequal prompt-engineering
effort would measure that effort rather than model capability.

Keyed, on-premises surrogate replacement lowers re-identification risk without
eliminating it \citep{sarkar2024deidnotenough}. It offers no formal privacy
guarantee of the kind defined by differential privacy \citep{dwork2006differential}.
The anonymizer checks in Table~\ref{tab:anon-metrics} are unit-level tests on synthetic
values; we did not run a linkage or re-identification attack on a release. We also did
not test whether surrogates make missed identifiers harder to find, as reported for
other systems \citep{carrell2013hiding}.
Membership inference is out of scope. We did not run the standard attack against a model
trained on our output \citep{shokri2017membership}, and we did not probe a released
encoder for memorised identifiers as has been done for clinical BERT models
\citep{lehman2021does}.

The recognizer stage is the component most likely to be superseded, and it was designed
to be replaceable. A site that adopts TIDE 2.0 with \sentryName{} can switch to a better
model later without re-anonymizing notes it has already processed. The crosswalk used
here to score i2b2 against a Stanford taxonomy can also be used to score a new model
against a site's own taxonomy. Shared evidence would also help. Useful contributions
include gazetteers of facility names and identifier formats, crosswalks between local
taxonomies, and per-category results in a comparable form. We release
\texttt{tide2-core} under an MIT license and the \sentryName{} recognizer at
\sentryRepo{} under a gated research-use agreement. We invite the OHDSI and open-source
clinical NLP communities to evaluate them and report their findings.

\section*{Methods}
\label{sec:methods}

\subsection*{Corpora}

\paragraph{SHIELD.}
SHIELD \citep{shield2025} comprises \shieldNotes{} Stanford Medicine clinical notes
annotated for \shieldNumCats{} PHI categories: AGE, DATE, DOCTOR, HOSPITAL, ID,
LOCATION, PATIENT, PHONE and WEB. The corpus totals \shieldGoldSpans{} gold spans. The
annotation protocol, inter-annotator agreement, and category definitions are given in
the corpus paper and are not re-derived here.

\paragraph{i2b2 2014.}
The i2b2 2014 shared-task corpus \citep{stubbs2015corpus} comprises longitudinal
records from a second institution, released with surrogate-replaced PHI. Because we
use it for evaluation only and train nothing on it, we pool all three track~1 releases,
the two training sets and the test set, giving \itwoNotes{} notes and \itwoRawSpans{}
annotated spans before crosswalking. Its identifiers are realistic surrogates rather than
real values, so recall on i2b2 may differ from recall on real identifiers from the same
institution.

\paragraph{Label crosswalk.}
The two corpora use different taxonomies, so i2b2 gold is crosswalked to the SHIELD
categories at scoring time. The mapping applies to the gold side only and does not
affect SHIELD scoring.
Four decisions define it. Each was measured on both corpora before adoption, so the
decisions were made with access to the evaluation data; Results report i2b2 recall
without the boundary alignment as well.

First, i2b2 DATE maps to SHIELD DATE unchanged. Philter's convention re-maps isolated
years, weekdays, seasons and bare month names as non-PHI, but SHIELD annotates isolated
years and weekdays too, and the affected forms are about a tenth of i2b2 DATE, so the
conventions agree closely enough to map straight through. The Philter re-mapping is
reported as a sensitivity analysis instead.

Second, i2b2 AGE maps to SHIELD AGE unchanged. HIPAA Safe Harbor treats only ages
of 90 and above as identifiers, but both corpora annotate ages at every value, and
SHIELD's AGE spans are dominated by values under 90, so the two conventions agree and
no re-mapping is required. The corpora do disagree on how much text an AGE span covers,
which is a boundary question handled below.

Third, the i2b2 \itwoExcludedTags{} tags are excluded from gold. Neither is a Safe
Harbor identifier and neither has a SHIELD counterpart. Any PROFESSION or ORGANIZATION
label that a recognizer emits maps to the unscored OTHER category, and neither
\sentryName{} nor the AIMI recognizers emit these labels, so removing the gold spans
creates no unmatched false negative and no unpenalised false positive for them. It leaves
one asymmetry for the external taggers: their
\texttt{company\_name} output maps to HOSPITAL, so a non-hospital organisation costs
them a false positive with no gold span to match.

Fourth, span boundaries are aligned. The two corpora annotate different extents for the
same identifier, which the overlap rule below would otherwise score as a detection
failure. i2b2 annotates the bare numeral of an age where SHIELD annotates the numeral
with its unit, so an i2b2 AGE span is extended over a following age unit
(\itwoAgeExtended{} spans extended). i2b2 annotates each address component separately
where SHIELD annotates the whole address, so adjacent spans that all map to LOCATION
and are separated by at most two characters of whitespace or punctuation are merged
(\itwoLocMerged{} merges). After the crosswalk, i2b2 gold is \itwoGoldSpans{} spans.
Supplementary Table~S6 gives the full mapping.

\paragraph{Leakage.}
The distilled recognizer was trained on teacher-generated labels over unannotated notes
that are disjoint from both corpora at the note level. We did not verify that their
patients are disjoint from SHIELD's. No gold annotation from either corpus entered its
training. The external taggers are public releases predating this evaluation. The AIMI
recognizers \citep{prakash_improving_2025} were also trained on i2b2 2014, so i2b2 is
in-domain for them and is not a held-out test set for those two models. Their training
data also include Stanford radiology reports, and we did not verify that these are
disjoint from SHIELD's notes or patients. Both corpora are therefore held out for the
distilled recognizer (SHIELD at the note level) and for the external taggers; neither is
verified as held out for the AIMI recognizers.

\subsection*{System architecture}

\begin{figure}[!htb]
  \centering
  \includegraphics[width=\textwidth,alt={TIDE 2.0 architecture. Clinical notes enter a Recognition Engine with four recognizer types: SLM transformers with NER BIO tagging, an LLM recognizer returning structured JSON spans, regex for phone, MRN, SSN, email and URL, and a gazetteer with Aho-Corasick lookup; all feed span resolution. An Anonymization Engine applies keyed selection for names and locations, format-preserving encryption for alphanumeric identifiers, per-patient date jitter, a SHA-256 accession hash, an age cap at 89 and faker replacement for email and URL, under a same input plus same keys equals same output rule, producing de-identified text written to Parquet, BigQuery or GCS. A Ray Data streaming band provides GPU/CPU actors, out-of-memory batch splitting and checkpointing. A cryptographic foundation provides key management, FF3 encryption, HMAC selection and date jitter. A properties box lists: deterministic per key, format-preserving, plausible surrogates, consistent across notes, and a new key gives an unlinkable release.}]{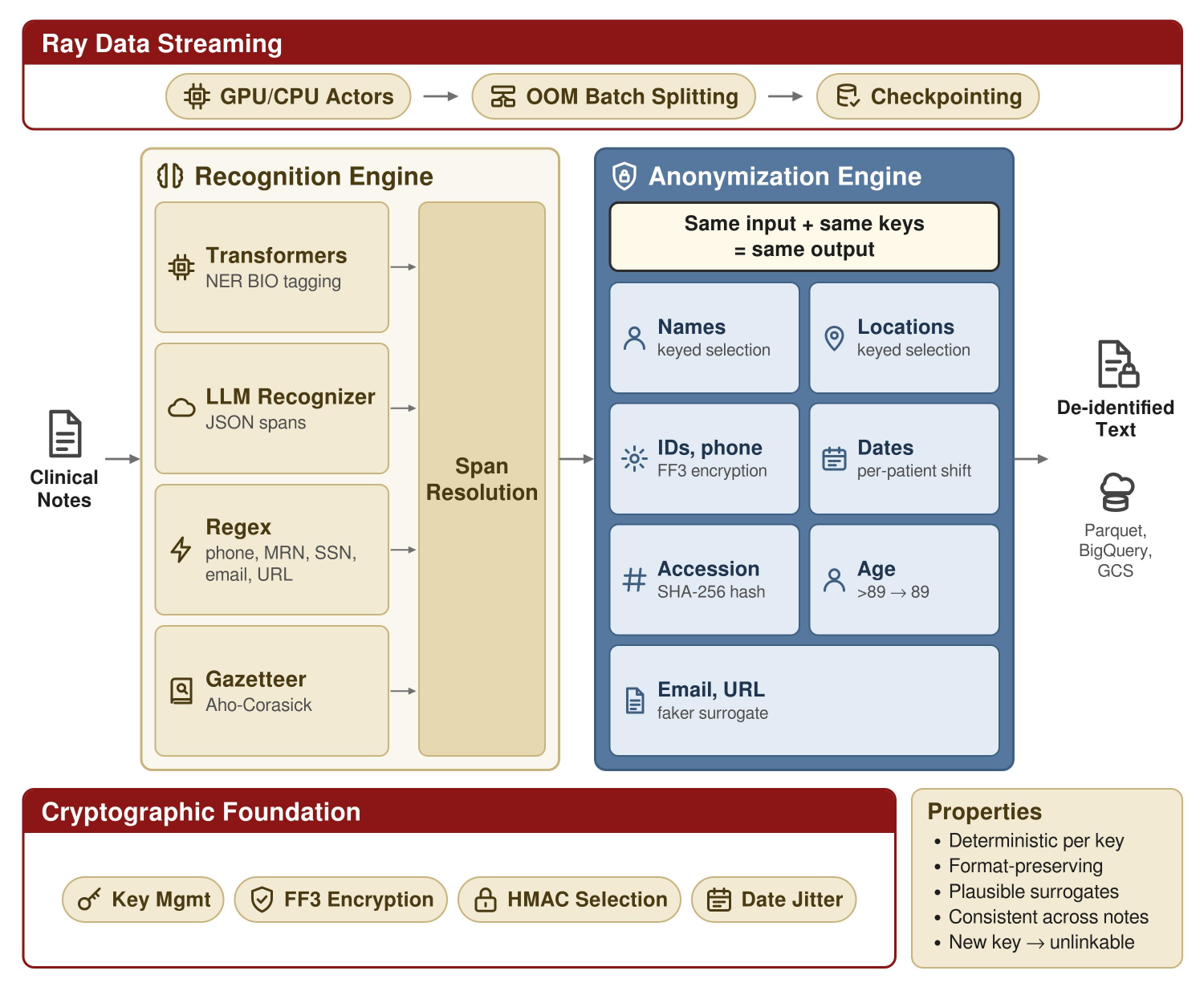}
  \caption{\textbf{Component architecture.} The recognizer slot accepts transformer
  token classifiers, an LLM recognizer, regular expressions and an Aho--Corasick
  gazetteer; span resolution passes a \texttt{RecognizerResult} list to the anonymizer,
  which applies keyed surrogates, format-preserving encryption, hashing and per-patient
  date jitter under a ``same input $+$ same keys $=$ same output'' rule. A Ray Data
  layer supplies GPU and CPU actors, out-of-memory batch splitting and checkpointing
  over a cryptographic foundation.}
  \label{fig:architecture}
\end{figure}

TIDE 2.0 (\texttt{tide2-core}) is an MIT-licensed Python~3.12 library. It separates a
\emph{recognizer stage}, which detects PHI spans, from an \emph{anonymizer stage}, which
replaces them (Figure~\ref{fig:pipeline}). The stages communicate through a cached span
representation, so a site can run, audit and re-run recognition without re-anonymizing,
and can change the strategy at either stage independently. Caching the spans is what
makes recognition auditable: a reviewer inspects what a model detected before any text
is altered. Figure~\ref{fig:architecture} shows the components described in the rest of
this section: the recognizer types and the span-resolution step, the per-category
anonymizer strategies, the Ray streaming layer, and the cryptographic foundation they
draw keys from.

\paragraph{Model-agnostic recognition.}
Recognition accepts four kinds of recognizer behind one interface: token classifiers,
an LLM recognizer, regular expressions and an Aho--Corasick gazetteer. Any token
classifier published through the Hugging Face \texttt{transformers} library
\citep{wolf2020transformers} registers through a Presidio-compatible interface
\citep{presidio2024} with no pipeline change. In this paper the slot held nine token
classifiers (\sentryName{}, the two AIMI recognizers and the \numExternalPII{} external
taggers) and the regex layer. An LLM-recognizer class prompts a
generative model to return PHI spans as structured JSON, placing cloud and open-weight
models in the same slot; here it was used only to run the distillation teacher. The same
interface exposes regular-expression and
Aho--Corasick matchers for structured identifiers such as medical record numbers,
encounter identifiers and social security numbers. Aho--Corasick matches an entire
gazetteer in a single pass over the note \citep{aho1975efficient}, so a site can add its
own facility names and identifier formats with no GPU work.

\paragraph{Distributed execution.}
Both stages run on Ray \citep{moritz2018ray} as a two-stage GPU/CPU pipeline. We chose Ray
over a general data-parallel engine because the job holds long-lived GPU model actors
and short CPU tasks in one topology, which the stage-and-executor model of a system such
as Spark \citep{zaharia2016spark} does not express directly. GPU actors
run model inference while concurrent CPU actors handle document windowing, reassembly,
regular-expression matching and anonymization. The GPU actor performs token-accurate
windowing internally, using a \windowBudget{}-token budget with \windowOverlap{}-token
overlap, so span offsets realign after reassembly. Batching is adaptive: a batch that
raises an out-of-memory error is split and retried rather than aborting the job.

\paragraph{Tooling.}
The library ships two command-line entry points: \texttt{tide2-runner}, which executes a
stage or a chained pipeline against an input corpus, and \texttt{tide2-visualizer}, a
Streamlit application that shows original and de-identified text side by side for human
verification. A warehouse ETL path writes the de-identified notes and their span
metadata back into the source schema. We describe this path but do not evaluate a
warehouse deployment in this paper.

\subsection*{Anonymizer and cryptographic construction}

The anonymizer replaces each detected span with a type-matched surrogate rather than a
redaction mark. Surrogates come from a keyed, table-free construction, so the same input
value maps to the same surrogate under a fixed key and to a different surrogate under a
rotated key, with no linkage database to store or protect. All primitives are standard.

\paragraph{Keys and salts.}
A salt is a 256-bit value from a cryptographically secure random source. A working key
is 32 bytes; where it derives from a study or tenant label it is
$\mathrm{SHA\text{-}256}(\texttt{label})$. Salt and key are combined as
$\texttt{salt}\,\|\,\texttt{key}$ before any derivation, so both are required to
reproduce a surrogate. Keys are isolated per study or tenant. A new key deterministically
re-maps every surrogate, so releases produced under different keys cannot be linked to
each other. Rotation does not alter releases already issued under the old key.

\paragraph{Per-patient date shift.}
The offset for a patient is
\[
  \mathrm{HMAC\text{-}SHA256}\bigl(\texttt{salt}\,\|\,\texttt{key},\;
  \texttt{``date\_jitter\_v1:''}\,\|\,\texttt{patient\_id}\bigr).
\]
HMAC is used rather than a bare hash of the concatenation because its
security does not depend on the key and the message being separable in the input, which
a plain $\mathrm{SHA\text{-}256}(\texttt{key}\,\|\,\texttt{message})$ construction does
depend on \citep{bellare1996keying}. The first four digest bytes
select the magnitude and a further byte the sign, giving an offset in
$[-180,-3]\cup[3,180]$ days; a minimum magnitude of three days is enforced, and the
domain-separation tag prevents cross-purpose reuse of the key. The same offset applies
to every date in that patient's record, which is what makes the shift
interval-preserving.

\paragraph{Numeric and alphanumeric identifiers.}
Identifiers are encrypted with FF3 format-preserving encryption over the digit
(radix~10) or alphanumeric (radix~62) alphabet. Format-preserving encryption is the
primitive that lets a medical record number be replaced by something a downstream system
will still accept as a medical record number \citep{bellare2009format}. FF3 was one of the
two modes NIST standardised for it in 2016 \citep{dworkin2016fpe}. The implementation
uses FF3 with a 64-bit tweak. The 128-bit cipher key is
$\mathrm{SHA\text{-}256}(\texttt{salt}\,\|\,\texttt{key})$ truncated to 16 bytes, and
the tweak derives per (format, content-length) as a keyed digest, so the same content
encrypts identically regardless of surrounding punctuation, which is reinserted
afterwards. Non-alphanumeric characters are preserved positionally, so length and
character class survive.

FF3 has documented attacks on small domains. A distinguishing and message-recovery
attack was published against the standardised mode \citep{durak2017breaking} and then
improved to full message recovery at practical cost \citep{hoang2019attacks}. NIST
responded with a revised mode, FF3-1, with a 56-bit tweak. A later analysis found a
weakness in the tweak schedule of both modes \citep{beyne2021linear}, and the current
draft revision of the NIST standard no longer specifies FF3 or FF3-1
\citep{dworkin2025fpe}. Both early attacks scale with the size of the domain, so a short
identifier is the dangerous case. The implementation therefore enforces a domain of at
least $10^6$: numeric content is padded to six digits and alphanumeric content to four
characters ($62^4 \approx 1.5\times10^6$) before encryption, then truncated back.
Truncation makes the map on short identifiers many-to-one, so distinct short identifiers
can share a surrogate (Results). Above FF3's maximum supported length (56 numeric, 32
alphanumeric characters) the construction falls back to a keyed $\mathrm{SHA\text{-}256}$.

\paragraph{Names, locations and opaque identifiers.}
Name and location surrogates use deterministic keyed selection: the index is
$\mathrm{SHA\text{-}256}(\texttt{salt}\,\|\,\texttt{key}\,\|\,\texttt{value}) \bmod |L|$
into a fixed surrogate list $L$, applied per token for multi-token names. Selection is
deterministic and one-way. State and country are retained for locations so region-level
covariates stay usable. Ages above 89 are set to 89, and ages at or below 89 are left
as written. Accession and other opaque identifiers are
replaced by
$\mathrm{SHA\text{-}256}(\texttt{salt}\,\|\,\texttt{study\_id}\,\|\,\texttt{entity}\,\|\,\texttt{id})$
truncated to 16 hex characters.

\subsection*{Threat model}

TIDE 2.0 produces real clinical text with synthetic PHI: the narrative is the original
note and only identifier spans are replaced, so the output is not a synthetic note in
the generative sense. It is also not a placeholder-substituted release of the kind
MIMIC-IV distributes, where the reader can see where each identifier used to be
\citep{johnson2023mimiciv}.

The design defends against a reader of a single release, in that name and opaque
identifier surrogates are one-way and the per-patient shift makes calendar
re-identification of a real date harder absent an external anchor event. It limits
linkage across releases, in that releases produced under different keys cannot be linked.
It does not protect releases already issued if their key is disclosed. It does not by
itself defeat membership inference or
re-identification from residual quasi-identifiers \citep{sarkar2024deidnotenough}.
Removing the named identifiers leaves combinations of age, date and location that can
still be distinguishing, which is the observation $k$-anonymity was introduced to
formalise \citep{sweeney2002kanonymity}. We make no $k$-anonymity claim about the
released text, and no claim that it meets the HIPAA Safe Harbor standard.

\subsection*{Distillation procedure}

The distilled recognizer, \sentryName{}, is a \sentryBase{} token classifier distilled
from \teacherName{} \citep{gemini25report}. The transfer is hard-label distillation
rather than the soft-target form \citep{hinton2015distilling}, because the teacher is
reached through an API that returns extracted spans and no logits. That is the usual
constraint when a language model is used to generate the training set for a much smaller
task model \citep{hsieh2023distilling}. The teacher was applied
to \teacherLabelNotes{} unannotated
clinical notes to produce silver-standard span annotations. These identified notes were
sent to the teacher through a cloud API under an executed business associate agreement;
no other step in this study sent identified text outside Stanford Medicine
infrastructure. The student was trained
on those labels with token-level cross-entropy over BIO tags across
\windowBudget{}-token windows. The student covers \sentryTypes{} label types: the
\shieldNumCats{} scored categories plus an \texttt{OTHER} catch-all for genetic
sequences, encoded image blobs and other miscellaneous identifiers, which the
evaluation does not score. The base encoder is DeBERTa-v3 \citep{he2021deberta};
long-context clinical encoders are alternatives for the same slot
\citep{sounack_bioclinical_2025,antoun_modernbert_2025}.

The teacher prompt is reproduced verbatim in Supplementary Note~1. It was developed on a
few dozen notes outside SHIELD and i2b2, and the teacher was then scored on all
\shieldNotes{} SHIELD notes to measure distillation fidelity.

\subsection*{Comparator recognizers}

Three fine-tuned clinical transformers were evaluated as drop-in recognizers: the two
AIMI radiology de-identification models \citep{prakash_improving_2025} and
\sentryName{}. The AIMI label set is DATE, HCW, HOSPITAL, ID, PATIENT, PHONE and VENDOR;
HCW maps to DOCTOR and VENDOR to HOSPITAL, and there is no AGE, LOCATION or WEB label.
TIDE 1.0 was run on SHIELD only. A regular expression and Aho--Corasick layer was
evaluated alone and as a
\texttt{+regex} configuration on each transformer; \sentryName{}\texttt{+regex} is the
default configuration. A further \numExternalPII{} external
open-source PII and clinical-NER taggers were evaluated on both corpora as an
out-of-distribution probe. They are trained on non-clinical or non-US identifier
distributions and are not tuned baselines.

Three of the external taggers were taken from the top five of a public PII-masking
leaderboard; two are smaller models from the same families, and one targets Indian
documents. A comparator had to load through the standard token-classification
pipeline without a separate integration path or a caller-supplied label prompt.
Supplementary Note~2 lists the models and the exclusions.

No large language model other than the teacher is evaluated. \teacherName{} was scored
on SHIELD only to measure distillation fidelity. The extraction prompt was developed for
this model alone, so a comparison across models would measure prompt fit rather than
model capability.

\subsection*{Throughput measurement}

Throughput was measured on \thruNumGpu{} \thruGpu{} GPU in a \thruPlatform{} with
\thruVcpus{} vCPUs and \thruRamGb{}~GB of RAM, using \sentryName{} with a
\windowBudget{}-token window budget and \windowOverlap{}-token overlap. We report
content-tokens per second, per stage and end-to-end, on two corpora: the full SHIELD
corpus (\thruShieldTokens{} content-tokens) and a bounded subset of the longest notes
(\thruLongTokens{} content-tokens) that exercises the long-note path. Content-tokens are
the recognizer tokenizer's tokens for the note text, excluding the model's special
tokens and counted once per note. The stages ran as one automatically chained job, and
each configuration was run once and timed end to end. This is a benchmark on one machine
rather than a measurement of a production deployment. The CPU-only comparison path uses
a single untuned actor, and GPU batch size was set separately for short and long notes.
Supplementary Note~3 gives the CPU actor sizing.

\subsection*{Outcomes and statistical analysis}

\paragraph{Declared outcomes.}
The primary outcome is span-level recall, per category and per corpus. Recall is primary
on an ethical rather
than a statistical rationale: a missed identifier is a disclosure with legal
consequence for a patient, whereas a false positive costs a clinical word. The two
errors are not interchangeable, so a single figure of merit that averages them, F1
included, cannot express the trade-off a site actually faces. Secondary outcomes are
span-level precision, which bounds how much clinical content survives replacement;
sustained throughput; and the anonymizer's structural guarantees.

\paragraph{Span matching.}
A predicted span $\hat{s}$ matches a gold span $s$ when they carry the same category and
their character offsets overlap by at least $\tau = \overlapThresh{}$ of the union. We
use a graded overlap criterion rather than the exact-match and partial-match pair
reported by the 2014 shared task \citep{stubbs2015overview}, because the two corpora
disagree on span extent often enough that exact match would score boundary conventions
instead of detection:
\begin{equation}
  \mathrm{IoU}(s, \hat{s}) \;=\; \frac{|s \cap \hat{s}|}{|s \cup \hat{s}|} \;\geq\; \tau .
\end{equation}
Matching runs per document, and each gold span matches at most one predicted span.
Writing $\mathrm{TP}$, $\mathrm{FP}$ and $\mathrm{FN}$ for the resulting counts,
\begin{equation}
  \mathrm{Recall} \;=\; \frac{\mathrm{TP}}{\mathrm{TP} + \mathrm{FN}},
  \qquad
  \mathrm{Precision} \;=\; \frac{\mathrm{TP}}{\mathrm{TP} + \mathrm{FP}} .
\end{equation}
Micro-averages pool $\mathrm{TP}$, $\mathrm{FP}$ and $\mathrm{FN}$ across categories;
macro-averages are the unweighted mean over the \shieldNumCats{} categories. As
sensitivity analyses we also report micro-averaged recall at $\tau = 0.5$, and
character-level recall: the share of gold PHI characters covered by any predicted span,
regardless of category.

\paragraph{Recall-drop decomposition.}
Micro-averaged recall is $R = \sum_c w_c r_c$, where $w_c$ is category $c$'s share of
gold spans and $r_c$ its recall. The drop from SHIELD ($s$) to i2b2 ($i$) splits exactly
into
\begin{equation}
  R_s - R_i \;=\; \sum_c \bar{w}_c\,(r_{s,c} - r_{i,c})
  \;+\; \sum_c \bar{r}_c\,(w_{s,c} - w_{i,c}),
\end{equation}
with $\bar{w}_c$ and $\bar{r}_c$ the means over the two corpora. The first sum is the
recall change per category, weighted by its share of gold spans; the second is the
effect of the different category mix.

\paragraph{Uncertainty and comparisons.}
Confidence intervals are 95\% bootstrap intervals over \bootstrapN{} document-level
resamples, resampling documents with replacement and recomputing the metric on each
resample \citep{efron1993bootstrap}. The document rather than the span is the resampling
unit because spans within one note are not independent. A facility acronym a model
cannot read tends to appear several times in the same note, and resampling spans would
count those as separate pieces of evidence. Teacher-student
recall differences are tested with a paired document-level bootstrap on the same
resamples, and $p$-values are compared against a Bonferroni threshold for a family of
\shieldNumCats{} categories ($0.05/\shieldNumCats{} \approx 0.006$), which is
conservative here because the per-category tests are positively correlated
\citep{dunn1961multiple}. We report Bonferroni-adjusted $p$-values, $\min(1, 9p)$. The
effect of the regex layer (\sentryName{}\texttt{+regex} minus \sentryName{}) on recall
and precision is estimated with the same paired document bootstrap, with 95\% percentile
intervals. The random seed is fixed at 42.

\section*{Data availability}

The SHIELD corpus is governed by its own release and is available under the terms
stated in the corpus paper \citep{shield2025}. The i2b2 2014 de-identification corpus
is available from the n2c2/DBMI data portal under the data use agreement described
there \citep{stubbs2015corpus}. The clinical notes used for distillation contain protected
health information and cannot be shared. Per-span predictions on both corpora are
available from the corresponding author, subject to the terms of each corpus.

\section*{Code availability}

\texttt{tide2-core}\footnote{\tidetworepo} is released under the MIT license. The
\sentryName{} recognizer is released on Hugging Face\footnote{\sentryRepo{}}, gated under Stanford Medicine's
Data Research Use Agreement (non-commercial research; no redistribution). This is a separate approval
from the SHIELD dataset.

\section*{Acknowledgements}
This research was funded, in part, by the Advanced Research Projects Agency for Health
(ARPA-H) under award number AY2AX000024. The views and conclusions contained in this
document are those of the authors and should not be interpreted as representing the
official policies, either expressed or implied, of the U.S. Government. The funders had
no role in study design, data collection and analysis, the decision to publish, or
preparation of the manuscript.

We gratefully acknowledge Professor Sylvia Plevritis, Ph.D. (Stanford University),
Principal Investigator of this project, for her support of this work. We also thank Todd
Ferris (Deputy Chief Information Officer, Stanford University School of Medicine) and
Anthea Buchin (Director of Research Technology Application Solutions, Stanford
Healthcare) for their leadership and institutional support. We thank our colleagues David Love, Joe
Pallas, Hannah Morgan-Cooper, Farnoosh Sheikhi, Shikha Kothari and Alvaro Alvarez, all of
Stanford Healthcare, for their contributions to the research data infrastructure and
engineering work that enabled this project.

\section*{Author contributions}
J.D.P.: Conceptualization, Methodology, Software, Investigation, Formal analysis, Writing.
S.D.: Conceptualization, Methodology, Writing -- review \& editing. 
P.D.: Conceptualization, Methodology, Writing -- review \& editing. 
All authors read and approved the final manuscript.

\section*{Competing interests}
The authors declare no competing interests.

\section*{Additional information}

\textbf{Supplementary Information}, with Supplementary Notes~1--3,
Fig.~S1 and Tables~S1--S14, follows the references.

\noindent\textbf{Correspondence} and requests for materials should be addressed to
J.D.P.

\bibliographystyle{unsrtnat}
\bibliography{references}

\clearpage
\appendix
\section*{Supplementary Information}
\setcounter{table}{0}
\setcounter{figure}{0}
\renewcommand{\thetable}{S\arabic{table}}
\renewcommand{\thefigure}{S\arabic{figure}}

\subsection*{Supplementary Note 1: teacher extraction prompt}

The teacher, \teacherName{}, was prompted to return PHI spans as JSON grouped by
category, giving for each span the exact surface text and a confidence score.
Character offsets are recovered by searching the note for each returned string, so a
span that differs from the note by one character is lost. This is why the prompt insists
on copying text exactly.

The prompt was developed on a few dozen notes from the same source as the distillation
set and disjoint from both evaluation corpora. The teacher was then scored once on all
\shieldNotes{} SHIELD notes to measure distillation fidelity. The prompt is reproduced
verbatim below. It is a Python format template: literal JSON braces appear doubled, and
\texttt{\{clinical\_text\}} is replaced by the note at call time.

\begin{Verbatim}

Identify and tag ALL patient-identifiable information and organizational information from clinical notes. Return results as JSON.

## TOP PRIORITIES (NEVER MISS THESE)

1. **DATES**: Scan exhaustively for ALL date formats (MM/DD/YYYY, M/D/YY, M/D, Month DD YYYY, etc.). Include short dates like "4/12" without years. Tag every occurrence separately.
- EXCLUDE times (10:00AM), temporal words (yesterday), day names (Monday). Example: "2023-03-05T03:43:00-08:00" -> "2023-03-05" (exclude T03:43:00-08:00).

2. **COMPLETE ADDRESSES WITH ZIP (CRITICAL PATTERN)**: ALWAYS tag addresses in format "Street Address  City State ZIP" (note: 2+ spaces between street and city). Example: "1234 MAIN STREET  BOSTON MA 02101" is ONE complete entity. This pattern appears frequently and MUST be captured.

3. **ALL DISTINCT MENTIONS**: Tag each occurrence separately:
   - Complete address: "5678 MAPLE STREET  BOSTON MA 02134" → tag it
   - Same street appearing alone: "5678 MAPLE STREET" → ALSO tag it separately when occurring in a different part of the text
   - Do NOT skip any mention just because another version exists elsewhere

4. **EXACT TEXT MATCHING**: Copy text exactly - preserve all case, whitespace, punctuation. Different whitespace = different entity ("Smith, John" vs "Smith,  John" are separate entities).

5. Check carefully for the entities separated by line breaks or several white spaces. Example: "Smith,  John   Albert"

## ENTITY DEFINITIONS

**PATIENT**: Patient names, relatives, friends, guardians, and any person name mentioned that is NOT a healthcare provider.
- EXCLUDE generic terms: "sister", "mother". EXCLUDE common professional titles that appear alongside the name. Example: John Smith PhD -> "John Smith" (exclude "PhD").
- EXCLUDE Ms./Mr./Mrs. Example: "Ms. Johnson" -> "Johnson" (exclude "Ms.").

**LOCATION**: ALL address mentions (complete or incomplete). Preserve internal whitespace exactly.
- Complete with ZIP: "456 OAK DR  Austin TX 78745" (one entity despite multiple spaces)
- Verbose format: "HealthMart #123 - Chicago, IL - 456 OAK DRIVE at Main St"
- Standalone street: "456 OAK DRIVE" (even if complete version exists elsewhere)
- Partial: "Chicago, IL - 456 OAK DRIVE"
- City/state: "Chicago, IL"
- Do NOT fragment single mentions. DO tag separate incomplete mentions. Do NOT double tag mentions that are already part of another mention.
- EXCLUDE generic facility names (PACU, ICU) without room numbers.

**DATE**: Any date format. Past, present, future.

**ID**: MRNs (include hyphens: "1234567-8"), SSNs, account numbers. EXCLUDE generic codes (R07.89), labels without values ("MRN").

**AGE**: Complete expressions with units ("72 Y old", "14 year old"). When units in separate fields ("Age: 14  Units: Years"), tag only number "14".
- EXCLUDE common female or male terms that accompany the age. Example 72 Y old female -> 72 Y old (exclude "female").

**PHONE**: All formats including pagers. For pagers, tag ONLY number ("12019"), NOT label ("Pgr").

**WEB**: Emails, URLs, social media handles, IPs.

**DOCTOR**: ALL healthcare worker names (providers, nurses, staff, interpreters, coordinators, financial staff, etc..):
- Formal: Examples:  "Dr. Smith", "Wilson, Anne Marie, MD"  -> "Smith", "Wilson, Anne Marie," (exclude "Dr.", "MD"), "Rn Johnson" -> "Johnson" (exclude "Rn")
- Informal: "spoke with Jennifer", "called Michael", "coordinator Tom"
- First names only when referring to healthcare workers
- If the title appears at the end, EXCLUDE it. Example: Smith, John Michael, MD -> "Smith, John Michael"
- If the title appears in the middle INCLUDE it. Example: Smith MD, John Michael Trey -> "Smith MD, John Michael Trey"

**HOSPITAL**: Organizations, pharmacies, clinics. Room numbers as number only ("3142", NOT "room 3142"). EXCLUDE generic teams (TT MICU GREEN), generic services (language line).

**OTHER**: Unique identifying characteristics.

## KEY RULES

**Organization + Location Combinations:**
- Full address: "HealthMart #7842 - Dallas, TX - 4825 Maple Dr" → "HealthMart #7842" (HOSPITAL) + ALL location variants (verbose full, ZIP format if present, standalone street, partial combinations)
- Shorthand (org + city/state only): "CVS Pharmacy Seattle, WA" → "CVS Pharmacy" (HOSPITAL) + "Seattle, WA" (LOCATION) as separate entities
- Multiple repetitions like this should be split into their components and tagged separately:
    - RITE AID-1991 MOUNTAIN BLVD. - OAKLAND, CA - 1991 MOUNTAIN BOULEVARD - > 1991 MOUNTAIN BLVD. - OAKLAND, CA (LOCATION) + 1991 MOUNTAIN BOULEVARD (LOCATION)
    - RITE AID-1991 MOUNTAIN BLVD. - OAKLAND, CA - 1991 MOUNTAIN BOULEVARD  1991 MOUNTAIN BOULEVARD  OAKLAND CA 94611-2812  -> 1991 MOUNTAIN BLVD. - OAKLAND, CA (LOCATION) + 1991 MOUNTAIN BOULEVARD (LOCATION) + 1991 MOUNTAIN BOULEVARD  OAKLAND CA 94611-2812 (LOCATION)

**Multiple Address Formats:** When same physical location appears in different formats, tag EACH format separately. Don't skip standard ZIP format just because verbose format appeared first.

**Exclusions:** Generic facilities (PACU), generic family terms (sister), generic teams (TT CARDIAC ICU), generic services (language line), medical codes (P245).

## OUTPUT FORMAT
Return valid JSON only. No markdown, explanations, or extra text.
```json
{{
  "ENTITY_TYPE": [
    {{"text": "exact text from input", "confidence": 0.95}},
    {{"text": "another entity", "confidence": 0.87}}
  ]
}}
```
Entity types: `PATIENT`, `LOCATION`, `PHONE`, `WEB`, `DATE`, `ID`, `HOSPITAL`, `DOCTOR`, `AGE`, `OTHER`
Confidence: 0.00-1.00 (use 0.90-0.99 for clear PHI)
No PHI found: return `{{}}`
Better to over-tag with lower confidence than miss entities.

## EXAMPLES

**Example 1 - Comprehensive Detection:**
Input: "Patient: Rodriguez, Maria Elena  MRN: 7834521-3  Age: 14  Age Units: Years  DOB: 03/22/2009
Pharmacy: MedPlus #5847 - Houston, TX - 2340 RIVERSIDE AVENUE at Memorial Drive
2340 RIVERSIDE AVENUE  Houston TX 77004
2340 RIVERSIDE AVENUE (standalone)
Phone: 713-555-8821  Fax: 713-555-8822  Pager Pgr 19304
Thomas K Williams is a 68 Y old male. Provider: Chen,  Lisa Marie, MD
Later, Dr. Chen, Lisa Marie (1 space) reviewed. Also consulted Smith, John Michael, MD.
Referred to Smith MD, John Michael Trey for cardiology. Ms. Johnson visited with family. Room 4217 at PACU.
Called Dr. Smith about results. Rn Johnson assisted with discharge.
Labs: HbA1c 07/15/2023, LIPID PANEL 01/08/2024, GLUCOSE  05/12/2024
Patient's brother visited. Team: TT NEURO ICU RED. Used language line. Code M79.3.
Spoke with Sandra in radiology. Interpreter Paulo #8472. Coordinator Michelle x6193.
Additional pharmacy: QuickCare Rx #2109 - Phoenix, AZ - 8765 DESERT BOULEVARD
8765 DESERT BOULEVARD  Phoenix AZ 85001
Verify: 8765 DESERT BOULEVARD
Prior address: 1542 OCEAN DRIVE  MIAMI FL 33139"

Output:
```json
{{
  "PATIENT": [
    {{"text": "Rodriguez, Maria Elena", "confidence": 0.95}},
    {{"text": "Thomas K Williams", "confidence": 0.95}},
    {{"text": "Johnson", "confidence": 0.93}}
  ],
  "ID": [
    {{"text": "7834521-3", "confidence": 0.99}},
    {{"text": "8472", "confidence": 0.94}}
  ],
  "AGE": [
    {{"text": "14", "confidence": 0.94}},
    {{"text": "68 Y old", "confidence": 0.94}}
  ],
  "DATE": [
    {{"text": "03/22/2009", "confidence": 0.98}},
    {{"text": "07/15/2023", "confidence": 0.98}},
    {{"text": "01/08/2024", "confidence": 0.98}},
    {{"text": "05/12/2024", "confidence": 0.98}}
  ],
  "HOSPITAL": [
    {{"text": "MedPlus #5847", "confidence": 0.95}},
    {{"text": "QuickCare Rx #2109", "confidence": 0.95}},
    {{"text": "4217", "confidence": 0.95}}
  ],
  "LOCATION": [
    {{"text": "Houston, TX - 2340 RIVERSIDE AVENUE at Memorial Drive", "confidence": 0.95}},
    {{"text": "2340 RIVERSIDE AVENUE  Houston TX 77004", "confidence": 0.97}},
    {{"text": "2340 RIVERSIDE AVENUE", "confidence": 0.94}},
    {{"text": "Phoenix, AZ - 8765 DESERT BOULEVARD", "confidence": 0.95}},
    {{"text": "8765 DESERT BOULEVARD  Phoenix AZ 85001", "confidence": 0.97}},
    {{"text": "8765 DESERT BOULEVARD", "confidence": 0.94}},
    {{"text": "1542 OCEAN DRIVE  MIAMI FL 33139", "confidence": 0.97}}
  ],
  "PHONE": [
    {{"text": "713-555-8821", "confidence": 0.98}},
    {{"text": "713-555-8822", "confidence": 0.98}},
    {{"text": "19304", "confidence": 0.94}},
    {{"text": "6193", "confidence": 0.94}}
  ],
  "DOCTOR": [
    {{"text": "Chen,  Lisa Marie,", "confidence": 0.93}},
    {{"text": "Chen, Lisa Marie", "confidence": 0.93}},
    {{"text": "Smith, John Michael,", "confidence": 0.93}},
    {{"text": "Smith MD, John Michael Trey", "confidence": 0.93}},
    {{"text": "Smith", "confidence": 0.93}},
    {{"text": "Johnson", "confidence": 0.92}},
    {{"text": "Sandra", "confidence": 0.91}},
    {{"text": "Paulo", "confidence": 0.91}},
    {{"text": "Michelle", "confidence": 0.91}}
  ]
}}
```

**Example 2 - Additional Patterns:**
Input: "At 0915 contacted Amy in billing x3482. Confirmed coverage.
Coordinator Jason arranged transport. Connected with interpreter Diana #7621.
Nurse Marcus reviewed discharge plan.
Primary pharmacy: CareMed #4783 - Boston, MA - 5123 HARBOR STREET
5123 HARBOR STREET  Boston MA 02108
5123 HARBOR STREET (confirm)
Alternate: WellRx Pharmacy Denver, CO (shorthand - no street)
Historical address: SupplyPlus - Atlanta, GA - 9876 PEACHTREE ROAD
9876 PEACHTREE ROAD  Atlanta GA 30303
Financial advisor Teresa x5247. Follow-up 11/5/2024. Email: patient@health.org"

Output:
```json
{{
  "DOCTOR": [
    {{"text": "Amy", "confidence": 0.91}},
    {{"text": "Jason", "confidence": 0.91}},
    {{"text": "Diana", "confidence": 0.91}},
    {{"text": "Marcus", "confidence": 0.91}},
    {{"text": "Teresa", "confidence": 0.91}}
  ],
  "PHONE": [
    {{"text": "3482", "confidence": 0.94}},
    {{"text": "7621", "confidence": 0.94}},
    {{"text": "5247", "confidence": 0.94}}
  ],
  "ID": [
    {{"text": "7621", "confidence": 0.94}}
  ],
  "HOSPITAL": [
    {{"text": "CareMed #4783", "confidence": 0.95}},
  ],
  "LOCATION": [
    {{"text": "Boston, MA - 5123 HARBOR STREET", "confidence": 0.96}},
    {{"text": "5123 HARBOR STREET  Boston MA 02108", "confidence": 0.97}},
    {{"text": "5123 HARBOR STREET", "confidence": 0.94}},
    {{"text": "Denver, CO", "confidence": 0.92}},
    {{"text": "Atlanta, GA - 9876 PEACHTREE ROAD", "confidence": 0.96}},
    {{"text": "9876 PEACHTREE ROAD  Atlanta GA 30303", "confidence": 0.97}}
  ],
  "DATE": [
    {{"text": "11/5/2024", "confidence": 0.97}}
  ],
  "WEB": [
    {{"text": "patient@health.org", "confidence": 0.98}}
  ]
}}
```

**Key Patterns Demonstrated:**
- MRNs with hyphens tagged
- Age with units vs separate field units handled correctly
- ALL dates captured (multiple years, different formats)
- **CRITICAL: Standard ZIP format "Street  City State ZIP" (2 spaces) ALWAYS captured** - appears 8 times across examples (including extended ZIP codes like 94301-2302 and concatenated ZIPs like 921284108)
- **Addresses with suite numbers, floor numbers, mail codes (MC), room/building codes (H310), and directional indicators (South) preserved exactly** - Examples: "4523 Maple Grove Suite 900", "850 Pine Avenue, 2nd Floor", "1250 Oak Boulevard MC 5807", "425 Forest Dr Mc 5208  H310", "9847 Sunset Creek Dr South Ste300"
- **State names can appear as abbreviations (CA) or full names (California)** - both formats captured
- Standalone street addresses tagged even when complete versions exist
- Verbose address formats captured
- Partial address combinations captured
- Doctor names with different whitespace variants captured separately
- **Doctor title handling: "Dr. Smith" → "Smith" (exclude "Dr."), "Wilson, Anne Marie, MD" → "Wilson, Anne Marie," (exclude "MD"), "Rn Johnson" → "Johnson" (exclude "Rn")**
- Informal healthcare worker first names tagged (billing, coordinators, interpreters, nurses, financial)
- Pager numbers without "Pgr" label
- Room numbers without "room" label
- Extension numbers as phone
- Shorthand pharmacy format split: "WellRx Pharmacy Denver, CO" → "Denver, CO" separately
- EXCLUDED: Generic facilities (PACU), generic terms (brother), generic teams (TT NEURO ICU RED), services (language line), medical codes (M79.3)

Clinical Text:

{clinical_text}
\end{Verbatim}

\subsection*{Supplementary Note 2: recognizer selection and exclusion}

The \numExternalPII{} external PII taggers are:
\begin{itemize}
  \item from the top five of a public PII-masking leaderboard (ai4privacy English F1):
  \path{lakshyakh93/deberta_finetuned_pii} (0.14\,B parameters),
  \path{OpenMed/OpenMed-PII-SuperClinical-Large-434M-v1} (0.43\,B) and
  \path{antoniogr7/pii-identifier-v0.1} (0.15\,B);
  \item two small encoders from families already represented, included to separate
  model size from training data:
  \path{OpenMed/OpenMed-PII-SuperClinical-Small-44M-v1} and
  \path{kalyan-ks/ettin-68m-nemotron-pii};
  \item the text model of \path{ekacare/document-pii-redactor}, which targets Indian
  names and documents and is an out-of-distribution check on US clinical notes.
\end{itemize}
The other two top-five leaderboard entries were excluded under the rule stated in
Methods, because they do not load through the standard token-classification pipeline.
The same rule excludes the GLiNER family, whose output depends on a caller-supplied
label prompt, and non-transformer models.

AIMI v1 and v2 are radiology-trained de-identification transformers
\citep{prakash_improving_2025} that were also trained on i2b2 2014. Their label set is
DATE, HCW, HOSPITAL, ID, PATIENT, PHONE and VENDOR (HCW scored as DOCTOR, VENDOR as
HOSPITAL); it has no AGE, LOCATION or WEB label.

\subsection*{Supplementary Note 3: Ray configuration and actor sizing}

Each CPU stage actor reserves one CPU and spawns a worker sub-actor that reserves
another. On the \thruVcpus{}~vCPU machine we used six actors, each with one CPU and a
one-CPU worker, plus half a CPU each for the reader and the writer: about 13 CPUs in
total. More actors over-subscribe the machine and deadlock the stage. The CPU-only
transformer path uses a single untuned actor. GPU batch size was set separately for short
and long notes.

\clearpage

\subsection*{Supplementary Figures}

\begin{figure}[H]
  \centering
  \includegraphics[width=0.9\textwidth,alt={Dumbbell plot of span-level precision and recall across nine PHI categories comparing the Gemini 2.5 Flash teacher and the TIDE2-Sentry student on SHIELD; red connectors indicate the student trails the teacher.}]{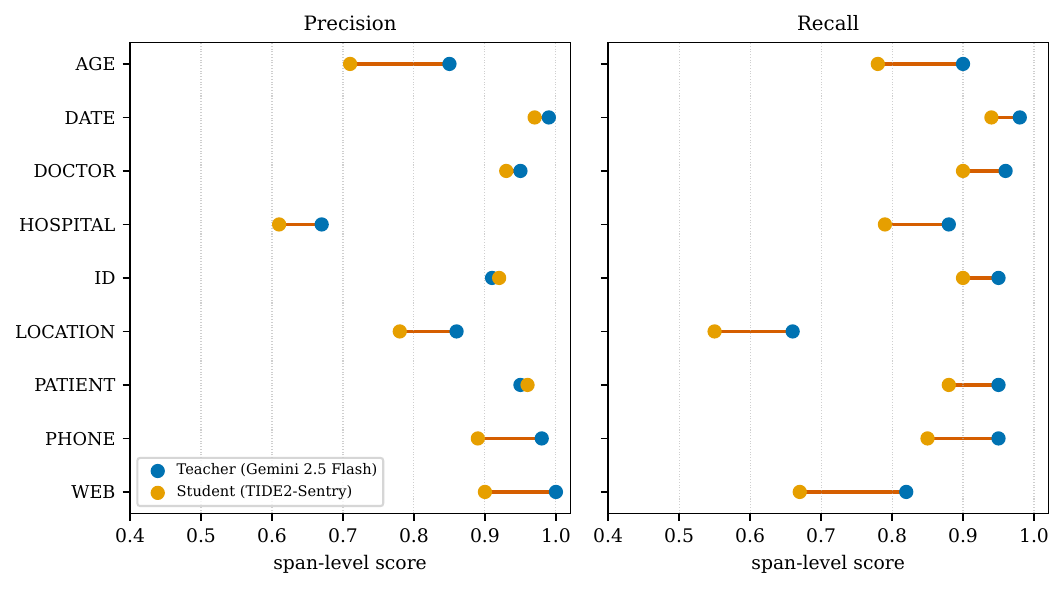}
  \caption{\textbf{Teacher and student per category on SHIELD.} Span-level precision
  (left) and recall (right) for the \teacherName{} teacher and the distilled
  \sentryName{} student. The student trails the teacher on recall in every category,
  most on WEB, AGE and LOCATION and least on DATE. Table~\ref{tab:distillation} gives the
  same comparison with confidence intervals and paired $p$-values.}
  \label{fig:distillation-dumbbell}
\end{figure}

\clearpage

\subsection*{Supplementary Tables}

\begin{table}[H]
  \caption{\textbf{Per-category span-level recall and precision on SHIELD.} Point
  estimates at an overlap threshold of \overlapThresh{} for regex, the two AIMI
  recognizers and \sentryName{} with their \texttt{+regex} ablations, and the
  \numExternalPII{} external PII taggers. ``--'' marks a category absent from the AIMI
  label set (AGE, LOCATION, WEB). The \texttt{+regex} rows isolate the effect of the
  regex layer. TIDE 1.0 is not shown; its SHIELD aggregates are in main-text Table~1.
  Every external tagger scores at or near zero on DOCTOR, because their taxonomies
  carry one generic person type where clinical de-identification needs the
  patient/provider distinction (pooled NAME results in Table~S14).}
  \label{tab:matrix-shield}
  \centering
  \scriptsize
  \resizebox{\textwidth}{!}{%
  \begin{tabular}{lccccccccc}
    \toprule
    \textbf{Model} & \textbf{AGE} & \textbf{DATE} & \textbf{DOCTOR} & \textbf{HOSP} & \textbf{ID} & \textbf{LOC} & \textbf{PATIENT} & \textbf{PHONE} & \textbf{WEB} \\
    \midrule
    \multicolumn{10}{l}{\emph{Recall}} \\
    TIDE2 regex & 0.00 & 0.92 & 0.00 & 0.00 & 0.56 & 0.38 & 0.00 & 0.80 & 0.82 \\
    AIMI v1 & -- & 0.95 & 0.46 & 0.90 & 0.95 & -- & 0.78 & 0.92 & -- \\
    AIMI v1 + regex & 0.00 & 0.95 & 0.46 & 0.90 & 0.95 & 0.36 & 0.78 & 0.94 & 0.60 \\
    AIMI v2 & -- & 0.95 & 0.42 & 0.89 & 0.95 & -- & 0.71 & 0.81 & -- \\
    AIMI v2 + regex & 0.00 & 0.96 & 0.42 & 0.88 & 0.95 & 0.37 & 0.71 & 0.86 & 0.56 \\
    TIDE2-Sentry & 0.78 & 0.94 & 0.90 & 0.79 & 0.90 & 0.55 & 0.88 & 0.85 & 0.67 \\
    TIDE2-Sentry + regex & 0.78 & 0.95 & 0.90 & 0.78 & 0.91 & 0.51 & 0.88 & 0.85 & 0.67 \\
    DeBERTa-PII (lakshyakh93) & 0.00 & 0.39 & 0.00 & 0.02 & 0.44 & 0.70 & 0.71 & 0.42 & 0.43 \\
    OpenMed SuperClinical L & 0.03 & 0.75 & 0.00 & 0.47 & 0.29 & 0.58 & 0.92 & 0.67 & 0.52 \\
    OpenMed SuperClinical S & 0.04 & 0.72 & 0.00 & 0.59 & 0.66 & 0.54 & 0.91 & 0.67 & 0.70 \\
    PII-Identifier (antonio) & 0.09 & 0.74 & 0.00 & 0.45 & 0.36 & 0.71 & 0.85 & 0.59 & 0.60 \\
    Ettin-68M Nemotron & 0.02 & 0.73 & 0.00 & 0.27 & 0.61 & 0.79 & 0.90 & 0.61 & 0.55 \\
    ekacare MiniLM & 0.88 & 0.66 & 0.00 & 0.00 & 0.62 & 0.72 & 0.81 & 0.81 & 0.73 \\
    \midrule
    \multicolumn{10}{l}{\emph{Precision}} \\
    TIDE2 regex & 0.00 & 0.98 & 0.00 & 0.00 & 0.94 & 0.93 & 0.00 & 0.98 & 0.82 \\
    AIMI v1 & -- & 0.94 & 0.41 & 0.47 & 0.63 & -- & 0.88 & 0.69 & -- \\
    AIMI v1 + regex & 0.00 & 0.93 & 0.42 & 0.53 & 0.72 & 0.93 & 0.89 & 0.87 & 0.77 \\
    AIMI v2 & -- & 0.95 & 0.37 & 0.39 & 0.52 & -- & 0.89 & 0.76 & -- \\
    AIMI v2 + regex & 0.00 & 0.94 & 0.38 & 0.44 & 0.61 & 0.93 & 0.90 & 0.92 & 0.75 \\
    TIDE2-Sentry & 0.71 & 0.97 & 0.93 & 0.61 & 0.92 & 0.78 & 0.96 & 0.89 & 0.90 \\
    TIDE2-Sentry + regex & 0.71 & 0.95 & 0.93 & 0.61 & 0.92 & 0.74 & 0.96 & 0.89 & 0.72 \\
    DeBERTa-PII (lakshyakh93) & 0.00 & 0.42 & 0.00 & 0.07 & 0.18 & 0.15 & 0.25 & 0.60 & 0.02 \\
    OpenMed SuperClinical L & 0.03 & 0.66 & 0.00 & 0.81 & 0.35 & 0.41 & 0.28 & 0.97 & 0.80 \\
    OpenMed SuperClinical S & 0.06 & 0.62 & 0.00 & 0.78 & 0.53 & 0.29 & 0.29 & 0.95 & 0.89 \\
    PII-Identifier (antonio) & 0.14 & 0.65 & 0.00 & 0.74 & 0.53 & 0.40 & 0.27 & 0.96 & 0.80 \\
    Ettin-68M Nemotron & 0.12 & 0.65 & 0.00 & 0.78 & 0.70 & 0.42 & 0.28 & 0.95 & 0.88 \\
    ekacare MiniLM & 0.72 & 0.57 & 0.00 & 0.00 & 0.32 & 0.42 & 0.22 & 0.91 & 0.90 \\
    \bottomrule
  \end{tabular}}
\end{table}

\begin{table}[H]
  \caption{\textbf{Per-category span-level recall and precision on i2b2 2014.} Same
  models and same conventions as Table~\ref{tab:matrix-shield}, after the label
  crosswalk of Table~\ref{tab:crosswalk}. The \numExternalPII{} external PII taggers
  are included as an off-distribution probe, as on SHIELD.}
  \label{tab:matrix-i2b2}
  \centering
  \scriptsize
  \resizebox{\textwidth}{!}{%
  \begin{tabular}{lccccccccc}
    \toprule
    \textbf{Model} & \textbf{AGE} & \textbf{DATE} & \textbf{DOCTOR} & \textbf{HOSP} & \textbf{ID} & \textbf{LOC} & \textbf{PATIENT} & \textbf{PHONE} & \textbf{WEB} \\
    \midrule
    \multicolumn{10}{l}{\emph{Recall}} \\
    TIDE2 regex & 0.00 & 0.75 & 0.00 & 0.00 & 0.44 & 0.13 & 0.00 & 0.38 & 0.71 \\
    AIMI v1 & -- & 0.94 & 0.25 & 0.87 & 0.81 & -- & 0.56 & 0.94 & -- \\
    AIMI v1 + regex & 0.00 & 0.95 & 0.25 & 0.87 & 0.81 & 0.08 & 0.56 & 0.93 & 0.57 \\
    AIMI v2 & -- & 0.94 & 0.23 & 0.84 & 0.80 & -- & 0.54 & 0.88 & -- \\
    AIMI v2 + regex & 0.00 & 0.95 & 0.23 & 0.84 & 0.80 & 0.09 & 0.54 & 0.88 & 0.29 \\
    TIDE2-Sentry & 0.68 & 0.85 & 0.85 & 0.39 & 0.64 & 0.65 & 0.79 & 0.56 & 0.57 \\
    TIDE2-Sentry + regex & 0.68 & 0.86 & 0.85 & 0.39 & 0.65 & 0.63 & 0.79 & 0.56 & 0.57 \\
    DeBERTa-PII (lakshyakh93) & 0.00 & 0.25 & 0.00 & 0.01 & 0.14 & 0.55 & 0.43 & 0.07 & 0.43 \\
    OpenMed SuperClinical L & 0.00 & 0.68 & 0.00 & 0.04 & 0.38 & 0.79 & 0.85 & 0.22 & 0.71 \\
    OpenMed SuperClinical S & 0.01 & 0.66 & 0.00 & 0.12 & 0.62 & 0.77 & 0.84 & 0.25 & 0.57 \\
    PII-Identifier (antonio) & 0.15 & 0.59 & 0.00 & 0.07 & 0.39 & 0.81 & 0.84 & 0.26 & 0.71 \\
    Ettin-68M Nemotron & 0.02 & 0.64 & 0.00 & 0.13 & 0.52 & 0.82 & 0.85 & 0.31 & 0.71 \\
    ekacare MiniLM & 0.75 & 0.70 & 0.00 & 0.00 & 0.44 & 0.81 & 0.49 & 0.22 & 0.71 \\
    \midrule
    \multicolumn{10}{l}{\emph{Precision}} \\
    TIDE2 regex & 0.00 & 0.99 & 0.00 & 0.00 & 0.91 & 0.85 & 0.00 & 0.97 & 0.62 \\
    AIMI v1 & -- & 0.98 & 0.25 & 0.59 & 0.83 & -- & 0.55 & 0.76 & -- \\
    AIMI v1 + regex & 0.00 & 0.98 & 0.25 & 0.61 & 0.85 & 0.84 & 0.55 & 0.82 & 0.57 \\
    AIMI v2 & -- & 0.98 & 0.23 & 0.52 & 0.73 & -- & 0.53 & 0.77 & -- \\
    AIMI v2 + regex & 0.15 & 0.97 & 0.23 & 0.54 & 0.76 & 0.85 & 0.53 & 0.83 & 0.40 \\
    TIDE2-Sentry & 0.74 & 0.96 & 0.92 & 0.48 & 0.85 & 0.69 & 0.90 & 0.73 & 0.22 \\
    TIDE2-Sentry + regex & 0.74 & 0.95 & 0.92 & 0.48 & 0.84 & 0.69 & 0.90 & 0.73 & 0.19 \\
    DeBERTa-PII (lakshyakh93) & 0.00 & 0.49 & 0.00 & 0.04 & 0.03 & 0.10 & 0.13 & 0.09 & 0.00 \\
    OpenMed SuperClinical L & 0.01 & 0.80 & 0.00 & 0.54 & 0.49 & 0.54 & 0.27 & 0.91 & 0.31 \\
    OpenMed SuperClinical S & 0.01 & 0.80 & 0.00 & 0.68 & 0.50 & 0.44 & 0.26 & 0.86 & 0.10 \\
    PII-Identifier (antonio) & 0.28 & 0.79 & 0.00 & 0.64 & 0.62 & 0.56 & 0.28 & 0.94 & 0.45 \\
    Ettin-68M Nemotron & 0.10 & 0.79 & 0.00 & 0.63 & 0.65 & 0.41 & 0.28 & 0.87 & 0.25 \\
    ekacare MiniLM & 0.70 & 0.70 & 0.00 & 0.00 & 0.34 & 0.48 & 0.16 & 0.44 & 0.20 \\
    \bottomrule
  \end{tabular}}
\end{table}

\begin{table}[H]
  \caption{\textbf{Distillation fidelity on SHIELD, per category.} \teacherName{}
  (teacher) against \sentryName{} (student), with bootstrap 95\% confidence intervals
  over \bootstrapN{} document-level resamples. $\Delta$R is student minus teacher
  recall, so a negative value means the student trails; $p$ comes from a paired
  document-bootstrap test, and $p_{\mathrm{adj}} = \min(1, \shieldNumCats{}\,p)$ is the
  Bonferroni-adjusted value for a family of \shieldNumCats{} categories.}
  \label{tab:distillation}
  \centering
  \small
  \resizebox{\textwidth}{!}{%
  \begin{tabular}{lccccrrr}
    \toprule
    & \multicolumn{2}{c}{\textbf{\teacherName{}}} & \multicolumn{2}{c}{\textbf{\sentryName{}}} & & & \\
    \cmidrule(lr){2-3} \cmidrule(lr){4-5}
    \textbf{Category} & \textbf{P [95\% CI]} & \textbf{R [95\% CI]} & \textbf{P [95\% CI]} & \textbf{R [95\% CI]} & \textbf{$\Delta$R} & \textbf{$p$} & \textbf{$p_{\mathrm{adj}}$} \\
    \midrule
    AGE & 0.85 [0.79-0.90] & 0.90 [0.87-0.93] & 0.71 [0.65-0.77] & 0.78 [0.74-0.83] & -0.12 & <0.0005 & <0.005 \\
    DATE & 0.99 [0.98-0.99] & 0.98 [0.98-0.99] & 0.97 [0.96-0.97] & 0.94 [0.93-0.95] & -0.04 & <0.0005 & <0.005 \\
    DOCTOR & 0.95 [0.94-0.97] & 0.96 [0.96-0.97] & 0.93 [0.92-0.94] & 0.90 [0.88-0.92] & -0.06 & <0.0005 & <0.005 \\
    HOSPITAL & 0.67 [0.64-0.71] & 0.88 [0.85-0.90] & 0.61 [0.57-0.65] & 0.79 [0.75-0.82] & -0.09 & <0.0005 & <0.005 \\
    ID & 0.91 [0.87-0.94] & 0.95 [0.92-0.97] & 0.92 [0.89-0.94] & 0.90 [0.86-0.94] & -0.05 & =0.0050 & 0.045 \\
    LOCATION & 0.86 [0.82-0.90] & 0.66 [0.60-0.71] & 0.78 [0.72-0.84] & 0.55 [0.49-0.61] & -0.11 & <0.0005 & <0.005 \\
    PATIENT & 0.95 [0.93-0.97] & 0.95 [0.93-0.97] & 0.96 [0.94-0.97] & 0.88 [0.85-0.90] & -0.07 & <0.0005 & <0.005 \\
    PHONE & 0.98 [0.97-0.99] & 0.95 [0.93-0.97] & 0.89 [0.85-0.92] & 0.85 [0.81-0.89] & -0.10 & <0.0005 & <0.005 \\
    WEB & 1.00 [1.00-1.00] & 0.82 [0.67-0.95] & 0.90 [0.81-0.98] & 0.67 [0.52-0.83] & -0.15 & =0.0225 & 0.202 \\
    \bottomrule
  \end{tabular}}
\end{table}

\begin{table}[H]
  \caption{\textbf{Per-category confidence intervals for the core recognizers.}
  Span-level precision and recall with bootstrap 95\% intervals over \bootstrapN{}
  document-level resamples, for regex, the fine-tuned clinical transformers and the
  \sentryName{}\texttt{+regex} ablation, on both corpora. The AIMI recognizers have no
  AGE, LOCATION or WEB label; their degenerate \mbox{0.00 [0.00--0.00]} cells in these
  categories reflect that, and Tables S1--S2 write them as ``--''.}
  \label{tab:core-ci}
  \centering
  \scriptsize
  \resizebox{\textwidth}{!}{%
  \begin{tabular}{llccccc}
    \toprule
    \textbf{Category} & \textbf{Metric} & \textbf{TIDE2 regex} & \textbf{AIMI v1} & \textbf{AIMI v2} & \textbf{\sentryName{}} & \textbf{\sentryName{}+regex} \\
    \midrule
    \multicolumn{7}{l}{\emph{SHIELD}} \\
    AGE & Precision & 0.00 [0.00-0.00] & 0.00 [0.00-0.00] & 0.00 [0.00-0.00] & 0.71 [0.65-0.77] & 0.71 [0.65-0.77] \\
    AGE & Recall & 0.00 [0.00-0.00] & 0.00 [0.00-0.00] & 0.00 [0.00-0.00] & 0.78 [0.74-0.83] & 0.78 [0.74-0.83] \\
    DATE & Precision & 0.98 [0.98-0.99] & 0.94 [0.93-0.95] & 0.95 [0.94-0.96] & 0.97 [0.96-0.97] & 0.95 [0.94-0.96] \\
    DATE & Recall & 0.92 [0.91-0.94] & 0.95 [0.94-0.96] & 0.95 [0.94-0.96] & 0.94 [0.93-0.95] & 0.95 [0.93-0.96] \\
    DOCTOR & Precision & 0.00 [0.00-0.00] & 0.41 [0.38-0.43] & 0.37 [0.34-0.39] & 0.93 [0.92-0.94] & 0.93 [0.92-0.94] \\
    DOCTOR & Recall & 0.00 [0.00-0.00] & 0.46 [0.43-0.49] & 0.42 [0.40-0.45] & 0.90 [0.88-0.92] & 0.90 [0.88-0.92] \\
    HOSPITAL & Precision & 0.00 [0.00-0.00] & 0.47 [0.44-0.50] & 0.39 [0.36-0.42] & 0.61 [0.57-0.65] & 0.61 [0.57-0.64] \\
    HOSPITAL & Recall & 0.00 [0.00-0.00] & 0.90 [0.88-0.93] & 0.89 [0.87-0.91] & 0.79 [0.75-0.82] & 0.78 [0.74-0.81] \\
    ID & Precision & 0.94 [0.91-0.97] & 0.63 [0.59-0.66] & 0.52 [0.49-0.56] & 0.92 [0.89-0.94] & 0.92 [0.89-0.94] \\
    ID & Recall & 0.56 [0.52-0.60] & 0.95 [0.92-0.98] & 0.95 [0.92-0.97] & 0.90 [0.86-0.94] & 0.91 [0.88-0.94] \\
    LOCATION & Precision & 0.93 [0.89-0.96] & 0.00 [0.00-0.00] & 0.00 [0.00-0.00] & 0.78 [0.72-0.84] & 0.74 [0.68-0.80] \\
    LOCATION & Recall & 0.38 [0.32-0.43] & 0.00 [0.00-0.00] & 0.00 [0.00-0.00] & 0.55 [0.49-0.61] & 0.51 [0.45-0.56] \\
    PATIENT & Precision & 0.00 [0.00-0.00] & 0.88 [0.86-0.91] & 0.89 [0.86-0.91] & 0.96 [0.94-0.97] & 0.96 [0.94-0.97] \\
    PATIENT & Recall & 0.00 [0.00-0.00] & 0.78 [0.75-0.81] & 0.71 [0.67-0.75] & 0.88 [0.85-0.90] & 0.88 [0.85-0.90] \\
    PHONE & Precision & 0.98 [0.96-1.00] & 0.69 [0.64-0.73] & 0.76 [0.72-0.80] & 0.89 [0.85-0.92] & 0.89 [0.85-0.92] \\
    PHONE & Recall & 0.80 [0.75-0.84] & 0.92 [0.89-0.94] & 0.81 [0.77-0.86] & 0.85 [0.81-0.89] & 0.85 [0.81-0.89] \\
    WEB & Precision & 0.82 [0.70-0.91] & 0.00 [0.00-0.00] & 0.00 [0.00-0.00] & 0.90 [0.81-0.98] & 0.72 [0.60-0.84] \\
    WEB & Recall & 0.82 [0.67-0.95] & 0.00 [0.00-0.00] & 0.00 [0.00-0.00] & 0.67 [0.52-0.83] & 0.67 [0.52-0.83] \\
    \midrule
    \multicolumn{7}{l}{\emph{i2b2 2014}} \\
    AGE & Precision & 0.00 [0.00-0.00] & 0.00 [0.00-0.00] & 0.14 [0.03-0.28] & 0.74 [0.72-0.77] & 0.74 [0.72-0.77] \\
    AGE & Recall & 0.00 [0.00-0.00] & 0.00 [0.00-0.00] & 0.00 [0.00-0.00] & 0.68 [0.66-0.71] & 0.68 [0.66-0.71] \\
    DATE & Precision & 0.99 [0.99-0.99] & 0.98 [0.98-0.99] & 0.98 [0.97-0.98] & 0.96 [0.95-0.96] & 0.95 [0.95-0.96] \\
    DATE & Recall & 0.75 [0.73-0.77] & 0.94 [0.93-0.95] & 0.94 [0.93-0.94] & 0.85 [0.84-0.86] & 0.86 [0.85-0.87] \\
    DOCTOR & Precision & 0.00 [0.00-0.00] & 0.25 [0.23-0.27] & 0.23 [0.21-0.25] & 0.92 [0.91-0.93] & 0.92 [0.91-0.93] \\
    DOCTOR & Recall & 0.00 [0.00-0.00] & 0.25 [0.23-0.27] & 0.23 [0.21-0.25] & 0.85 [0.84-0.87] & 0.85 [0.84-0.87] \\
    HOSPITAL & Precision & 0.00 [0.00-0.00] & 0.59 [0.57-0.61] & 0.52 [0.49-0.54] & 0.48 [0.45-0.51] & 0.48 [0.45-0.51] \\
    HOSPITAL & Recall & 0.00 [0.00-0.00] & 0.87 [0.85-0.88] & 0.84 [0.82-0.86] & 0.39 [0.37-0.42] & 0.39 [0.37-0.42] \\
    ID & Precision & 0.91 [0.87-0.94] & 0.83 [0.81-0.85] & 0.73 [0.71-0.75] & 0.85 [0.82-0.87] & 0.84 [0.82-0.86] \\
    ID & Recall & 0.44 [0.40-0.48] & 0.81 [0.75-0.87] & 0.80 [0.74-0.86] & 0.64 [0.58-0.69] & 0.65 [0.59-0.70] \\
    LOCATION & Precision & 0.85 [0.79-0.91] & 0.00 [0.00-0.00] & 0.00 [0.00-0.00] & 0.69 [0.65-0.73] & 0.69 [0.65-0.72] \\
    LOCATION & Recall & 0.13 [0.11-0.16] & 0.00 [0.00-0.00] & 0.00 [0.00-0.00] & 0.65 [0.61-0.69] & 0.63 [0.59-0.67] \\
    PATIENT & Precision & 0.00 [0.00-0.00] & 0.55 [0.52-0.57] & 0.53 [0.50-0.56] & 0.90 [0.88-0.93] & 0.90 [0.88-0.93] \\
    PATIENT & Recall & 0.00 [0.00-0.00] & 0.56 [0.53-0.59] & 0.54 [0.51-0.57] & 0.79 [0.77-0.82] & 0.79 [0.77-0.82] \\
    PHONE & Precision & 0.97 [0.94-1.00] & 0.76 [0.72-0.80] & 0.77 [0.73-0.81] & 0.73 [0.68-0.77] & 0.73 [0.68-0.78] \\
    PHONE & Recall & 0.38 [0.33-0.42] & 0.94 [0.92-0.96] & 0.88 [0.85-0.91] & 0.56 [0.52-0.60] & 0.56 [0.51-0.60] \\
    WEB & Precision & 0.62 [0.25-1.00] & 0.00 [0.00-0.00] & 0.00 [0.00-0.00] & 0.22 [0.04-0.47] & 0.19 [0.04-0.40] \\
    WEB & Recall & 0.71 [0.25-1.00] & 0.00 [0.00-0.00] & 0.00 [0.00-0.00] & 0.57 [0.10-1.00] & 0.57 [0.10-1.00] \\
    \bottomrule
  \end{tabular}}
\end{table}

\begin{table}[H]
  \caption{\textbf{Per-category confidence intervals for the external PII taggers.}
  Bootstrap 95\% intervals over \bootstrapN{} document-level resamples, on SHIELD and
  i2b2 2014. These models are trained on non-clinical or non-US PII distributions, so
  their numbers read as an off-distribution check and not as tuned baselines
  (Supplementary Note 2).}
  \label{tab:ext-ci}
  \centering
  \scriptsize
  \resizebox{\textwidth}{!}{%
  \begin{tabular}{llcccccc}
    \toprule
    \textbf{Category} & \textbf{Metric} & \textbf{DeBERTa-PII} & \textbf{OpenMed L} & \textbf{OpenMed S} & \textbf{PII-Ident.} & \textbf{Ettin-68M} & \textbf{ekacare} \\
    \midrule
    \multicolumn{8}{l}{\emph{SHIELD}} \\
    AGE & Precision & 0.00 [0.00-0.00] & 0.03 [0.01-0.06] & 0.06 [0.04-0.09] & 0.14 [0.09-0.18] & 0.12 [0.05-0.20] & 0.72 [0.68-0.77] \\
    AGE & Recall & 0.00 [0.00-0.00] & 0.03 [0.01-0.05] & 0.04 [0.03-0.07] & 0.09 [0.06-0.11] & 0.02 [0.01-0.04] & 0.88 [0.85-0.92] \\
    DATE & Precision & 0.42 [0.40-0.44] & 0.66 [0.64-0.68] & 0.62 [0.59-0.64] & 0.65 [0.63-0.67] & 0.65 [0.63-0.67] & 0.57 [0.54-0.59] \\
    DATE & Recall & 0.39 [0.36-0.42] & 0.75 [0.73-0.77] & 0.72 [0.70-0.75] & 0.74 [0.71-0.76] & 0.73 [0.71-0.75] & 0.66 [0.63-0.68] \\
    DOCTOR & Precision & 0.00 [0.00-0.00] & 0.00 [0.00-0.00] & 0.00 [0.00-0.00] & 0.00 [0.00-0.00] & 0.00 [0.00-0.00] & 0.00 [0.00-0.00] \\
    DOCTOR & Recall & 0.00 [0.00-0.00] & 0.00 [0.00-0.00] & 0.00 [0.00-0.00] & 0.00 [0.00-0.00] & 0.00 [0.00-0.00] & 0.00 [0.00-0.00] \\
    HOSPITAL & Precision & 0.07 [0.04-0.11] & 0.81 [0.77-0.85] & 0.78 [0.74-0.82] & 0.74 [0.69-0.78] & 0.78 [0.72-0.83] & 0.00 [0.00-0.00] \\
    HOSPITAL & Recall & 0.02 [0.01-0.03] & 0.47 [0.43-0.50] & 0.59 [0.55-0.62] & 0.45 [0.41-0.49] & 0.27 [0.24-0.30] & 0.00 [0.00-0.00] \\
    ID & Precision & 0.18 [0.16-0.20] & 0.35 [0.31-0.38] & 0.53 [0.50-0.56] & 0.53 [0.49-0.58] & 0.70 [0.66-0.74] & 0.32 [0.29-0.35] \\
    ID & Recall & 0.44 [0.39-0.48] & 0.29 [0.25-0.33] & 0.66 [0.62-0.70] & 0.36 [0.32-0.41] & 0.61 [0.56-0.66] & 0.62 [0.58-0.66] \\
    LOCATION & Precision & 0.15 [0.13-0.17] & 0.41 [0.36-0.46] & 0.29 [0.25-0.34] & 0.40 [0.35-0.45] & 0.42 [0.37-0.46] & 0.42 [0.38-0.47] \\
    LOCATION & Recall & 0.70 [0.65-0.75] & 0.58 [0.52-0.63] & 0.54 [0.48-0.60] & 0.71 [0.66-0.76] & 0.79 [0.74-0.83] & 0.72 [0.67-0.77] \\
    PATIENT & Precision & 0.25 [0.23-0.28] & 0.28 [0.26-0.30] & 0.29 [0.27-0.32] & 0.27 [0.25-0.29] & 0.28 [0.26-0.30] & 0.22 [0.20-0.24] \\
    PATIENT & Recall & 0.71 [0.68-0.75] & 0.92 [0.90-0.93] & 0.91 [0.89-0.93] & 0.85 [0.82-0.87] & 0.90 [0.88-0.92] & 0.81 [0.78-0.84] \\
    PHONE & Precision & 0.60 [0.54-0.66] & 0.97 [0.95-0.99] & 0.95 [0.92-0.97] & 0.96 [0.94-0.98] & 0.95 [0.92-0.97] & 0.91 [0.86-0.94] \\
    PHONE & Recall & 0.42 [0.37-0.47] & 0.67 [0.62-0.72] & 0.67 [0.62-0.72] & 0.59 [0.54-0.64] & 0.61 [0.55-0.67] & 0.81 [0.76-0.85] \\
    WEB & Precision & 0.02 [0.01-0.03] & 0.80 [0.65-0.92] & 0.89 [0.81-0.96] & 0.80 [0.70-0.89] & 0.88 [0.77-0.96] & 0.90 [0.81-0.97] \\
    WEB & Recall & 0.43 [0.29-0.57] & 0.52 [0.38-0.69] & 0.70 [0.54-0.84] & 0.60 [0.45-0.75] & 0.55 [0.41-0.69] & 0.73 [0.57-0.89] \\
    \midrule
    \multicolumn{8}{l}{\emph{i2b2 2014}} \\
    AGE & Precision & 0.00 [0.00-0.00] & 0.01 [0.00-0.01] & 0.01 [0.00-0.01] & 0.28 [0.26-0.31] & 0.10 [0.07-0.14] & 0.70 [0.68-0.72] \\
    AGE & Recall & 0.00 [0.00-0.00] & 0.00 [0.00-0.01] & 0.01 [0.00-0.01] & 0.15 [0.14-0.17] & 0.02 [0.01-0.02] & 0.75 [0.72-0.77] \\
    DATE & Precision & 0.49 [0.47-0.51] & 0.80 [0.77-0.83] & 0.80 [0.76-0.82] & 0.79 [0.75-0.83] & 0.79 [0.75-0.82] & 0.70 [0.67-0.72] \\
    DATE & Recall & 0.25 [0.24-0.27] & 0.68 [0.66-0.71] & 0.66 [0.64-0.68] & 0.59 [0.56-0.61] & 0.64 [0.61-0.67] & 0.70 [0.67-0.72] \\
    DOCTOR & Precision & 0.00 [0.00-0.00] & 0.00 [0.00-0.00] & 0.00 [0.00-0.00] & 0.00 [0.00-0.00] & 0.00 [0.00-0.00] & 0.00 [0.00-0.00] \\
    DOCTOR & Recall & 0.00 [0.00-0.00] & 0.00 [0.00-0.00] & 0.00 [0.00-0.00] & 0.00 [0.00-0.00] & 0.00 [0.00-0.00] & 0.00 [0.00-0.00] \\
    HOSPITAL & Precision & 0.04 [0.02-0.05] & 0.54 [0.46-0.62] & 0.68 [0.63-0.73] & 0.64 [0.57-0.70] & 0.63 [0.58-0.67] & 0.00 [0.00-0.00] \\
    HOSPITAL & Recall & 0.01 [0.01-0.02] & 0.04 [0.03-0.05] & 0.12 [0.10-0.14] & 0.07 [0.06-0.08] & 0.13 [0.11-0.14] & 0.00 [0.00-0.00] \\
    ID & Precision & 0.03 [0.02-0.03] & 0.49 [0.44-0.54] & 0.50 [0.47-0.52] & 0.62 [0.57-0.66] & 0.65 [0.61-0.69] & 0.34 [0.31-0.36] \\
    ID & Recall & 0.14 [0.12-0.16] & 0.38 [0.34-0.41] & 0.62 [0.58-0.66] & 0.39 [0.36-0.42] & 0.52 [0.48-0.56] & 0.44 [0.40-0.48] \\
    LOCATION & Precision & 0.10 [0.09-0.11] & 0.54 [0.52-0.57] & 0.44 [0.42-0.47] & 0.56 [0.53-0.58] & 0.41 [0.39-0.43] & 0.48 [0.45-0.50] \\
    LOCATION & Recall & 0.55 [0.51-0.59] & 0.79 [0.76-0.82] & 0.77 [0.74-0.80] & 0.81 [0.78-0.84] & 0.82 [0.79-0.84] & 0.81 [0.78-0.84] \\
    PATIENT & Precision & 0.13 [0.12-0.14] & 0.27 [0.25-0.28] & 0.26 [0.25-0.28] & 0.28 [0.26-0.29] & 0.28 [0.27-0.30] & 0.16 [0.15-0.17] \\
    PATIENT & Recall & 0.43 [0.40-0.45] & 0.85 [0.83-0.86] & 0.84 [0.82-0.86] & 0.84 [0.82-0.86] & 0.85 [0.83-0.87] & 0.49 [0.46-0.52] \\
    PHONE & Precision & 0.09 [0.06-0.13] & 0.91 [0.85-0.95] & 0.86 [0.80-0.91] & 0.94 [0.90-0.97] & 0.87 [0.80-0.92] & 0.44 [0.37-0.51] \\
    PHONE & Recall & 0.07 [0.04-0.10] & 0.22 [0.18-0.25] & 0.25 [0.21-0.29] & 0.26 [0.22-0.30] & 0.31 [0.27-0.35] & 0.22 [0.18-0.26] \\
    WEB & Precision & 0.00 [0.00-0.00] & 0.31 [0.09-0.54] & 0.10 [0.02-0.21] & 0.45 [0.14-0.75] & 0.25 [0.06-0.46] & 0.20 [0.05-0.38] \\
    WEB & Recall & 0.43 [0.00-1.00] & 0.71 [0.25-1.00] & 0.57 [0.10-1.00] & 0.71 [0.25-1.00] & 0.71 [0.25-1.00] & 0.71 [0.25-1.00] \\
    \bottomrule
  \end{tabular}}
\end{table}

\begin{table}[H]
  \caption{\textbf{The i2b2 2014 to SHIELD label crosswalk.} The mapping is applied to
  the gold spans at scoring time. After exclusions and boundary alignment the corpus has
  \itwoGoldSpans{} gold spans over \itwoNotes{} notes. The boundary rules follow from the
  annotation conventions: i2b2 annotates most ages as a bare number and each address
  component separately, whereas SHIELD includes the age unit and annotates the whole
  address as one span.}
  \label{tab:crosswalk}
  \centering
  \small
  \begin{tabular}{P{0.144\linewidth}P{0.312\linewidth}P{0.449\linewidth}}
    \toprule
    \textbf{SHIELD category} & \textbf{i2b2 2014 tags} & \textbf{Treatment} \\
    \midrule
    AGE & AGE & straight through; right edge extended over the age unit \\
    DATE & DATE & straight through (Philter safe-date re-mapping: sensitivity only) \\
    DOCTOR & DOCTOR & straight through \\
    HOSPITAL & HOSPITAL & straight through \\
    ID & MEDICALRECORD /\allowbreak{} IDNUM /\allowbreak{} USERNAME /\allowbreak{} DEVICE /\allowbreak{} HEALTHPLAN /\allowbreak{} BIOID & relabelled \\
    LOCATION & CITY /\allowbreak{} STATE /\allowbreak{} STREET /\allowbreak{} ZIP /\allowbreak{} COUNTRY /\allowbreak{} LOCATION-OTHER & relabelled; adjacent components merged into one span \\
    PATIENT & PATIENT & straight through \\
    PHONE & PHONE /\allowbreak{} FAX & relabelled \\
    WEB & EMAIL /\allowbreak{} URL & relabelled \\
    (not scored) & ORGANIZATION & excluded from gold; model output already unscored \\
    (not scored) & PROFESSION & excluded from gold; model output already unscored \\
    \bottomrule
  \end{tabular}
\end{table}

\begin{table}[H]
  \caption{\textbf{Functional and architectural change from TIDE 1.0 to TIDE 2.0.}}
  \label{tab:tide-capabilities}
  \centering
  \small
  \begin{tabular}{P{0.271\linewidth}P{0.271\linewidth}P{0.362\linewidth}}
    \toprule
    \textbf{Capability} & \textbf{TIDE 1.0} & \textbf{TIDE 2.0} \\
    \midrule
    Recognizer & coupled\tblcomma{} fixed NER & model-agnostic (regex /\allowbreak{} transformer /\allowbreak{} LLM) \\
    Name /\allowbreak{} location surrogate & yes (RNG\tblcomma{} non-deterministic) & yes (cryptographic keyed) \\
    Same value $\to$ same surrogate & no & yes\tblcomma{} automatic (keyed) \\
    New key gives an unlinkable release & n/\allowbreak{}a & yes \\
    ID /\allowbreak{} MRN /\allowbreak{} SSN /\allowbreak{} phone & redaction only & format-preserving encryption /\allowbreak{} hash \\
    Email /\allowbreak{} URL /\allowbreak{} organization & redaction only & surrogate \\
    Age handling & generalize 90+ only & cap at 89 \\
    State /\allowbreak{} country retained & yes & yes \\
    Distributed execution & no & Ray GPU/\allowbreak{}CPU\tblcomma{} adaptive batching \\
    Out-of-memory recovery & no & automatic batch-splitting \\
    License & open source & MIT \\
    \bottomrule
  \end{tabular}
\end{table}

\begin{table}[H]
  \caption{\textbf{Worked anonymization examples on identical inputs.} TIDE 1.0 redacts
  structured identifiers and draws non-deterministic name and location surrogates;
  TIDE 2.0 applies a surrogate, format-preserving encryption or a keyed hash to every
  category.}
  \label{tab:anon-examples}
  \centering
  \small
  \resizebox{\textwidth}{!}{%
  \begin{tabular}{l l P{2.8cm} P{1.9cm} P{4.4cm} P{1.9cm}}
    \toprule
    \textbf{Category} & \textbf{PHI span} & \textbf{TIDE 1.0} & \textbf{TIDE 1.0 strat.} & \textbf{TIDE 2.0} & \textbf{TIDE 2.0 strat.} \\
    \midrule
    Name & John Smith & Ziyad Bromberg & surrogate (RNG) & Flynn Singleton & keyed surrogate \\
    Location & 123 Main St\tblcomma{} New York\tblcomma{} NY 10001 & Nashua NH 03060 & surrogate (RNG) & 13737 Chestnut Hill Rd\tblcomma{} Balance Of Jerome County\tblcomma{} NY 49885 & keyed surrogate \\
    Org & Stanford Health Care & Stanford Health Care & unchanged (no org surrogate) & Unity Ridge Community Hospital & keyed surrogate \\
    Date & April 18\tblcomma{} 2012 & 04/28/2012 & date-shift (jitter) & April 28th\tblcomma{} 2012 & date-shift (jitter) \\
    Age & 95 years old & [90 year old] old & generalization (90+) & 89 years old & cap at 89 \\
    Age & 88 years old & 88 years old & unchanged (fires 90+ only) & 88 years old & cap at 89 \\
    MRN & 15851368 & [MRN] & redaction & 15068153 & FPE (FF3) \\
    SSN & 123-45-6789 & [general-ssn] & redaction & 069-88-6168 & FPE (FF3) \\
    Phone & 650-876-9087 & [general-phone] & redaction & 504-015-2308 & FPE (FF3) \\
    Account & 22222222 & [general-account] & redaction & 743A4DE6F3E9B49A & hash (SHA256) \\
    Email & wcl@com.com & [general-email] & redaction & johnsonjoshua@example.org & faker surrogate \\
    URL & https:/\allowbreak{}/\allowbreak{}www.yahoo.com/\allowbreak{}testpage & [general-url] & redaction & https:/\allowbreak{}/\allowbreak{}www.hill.net/\allowbreak{} & faker surrogate \\
    \bottomrule
  \end{tabular}}
\end{table}

\begin{table}[H]
  \caption{\textbf{Consistency and remapping.} Each value appears three times across
  two notes for one patient. TIDE 2.0 maps a value to the same surrogate under a fixed
  key, which is what makes cross-note linkage survive, and to a different surrogate
  under a new key, so releases under different keys cannot be linked. TIDE 1.0 draws a new
  random surrogate at each occurrence for names.}
  \label{tab:anon-consistency}
  \centering
  \small
  \begin{tabular}{lllll}
    \toprule
    \textbf{Value} & \textbf{Occ.} & \textbf{TIDE 1.0} & \textbf{TIDE 2.0 (same key)} & \textbf{TIDE 2.0 (new key)} \\
    \midrule
    Jose (Name) & A/1 & Rourke & Tidwell & Rowland \\
     & A/2 & Suskin & Tidwell & (same) \\
     & B/1 & Abid & Tidwell & (same) \\
    15851368 (MRN) & A/1 & [MRN] & 15068153 & 24176984 \\
     & A/2 & [MRN] & 15068153 & (same) \\
     & B/1 & [MRN] & 15068153 & (same) \\
    03/15/2023 (Date) & A/1 & 03/25/2023 & 03/25/2023 & 03/25/2023 \\
     & A/2 & 03/25/2023 & 03/25/2023 & (same) \\
     & B/1 & 03/25/2023 & 03/25/2023 & (same) \\
    \bottomrule
  \end{tabular}
\end{table}

\begin{table}[H]
  \caption{\textbf{Feature comparison with OpenMed, a per-document detection toolkit.}
  The two systems address different problems and are complementary: TIDE 2.0 is a
  batch engine and can host an OpenMed model as a recognizer plugin.}
  \label{tab:openmed}
  \centering
  \small
  \begin{tabular}{P{0.243\linewidth}P{0.351\linewidth}P{0.310\linewidth}}
    \toprule
    \textbf{Axis} & \textbf{TIDE 2.0 (batch engine)} & \textbf{OpenMed (detection toolkit)} \\
    \midrule
    Primary artifact & de-id engine/\allowbreak{}pipeline (recognizer-agnostic) & model collection + toolkit \\
    Target unit of work & batch corpora & per-document /\allowbreak{} on-device inference \\
    Recognizer flexibility & any HF token classifier\tblcomma{} LLM\tblcomma{} regex/\allowbreak{}Aho-Corasick & own MoE Privacy-Filter + NER families \\
    LLM recognizer support & yes (structured-JSON extraction) & not provided \\
    Anonymization strategies & mask /\allowbreak{} keyed surrogate /\allowbreak{} FPE /\allowbreak{} hash /\allowbreak{} per-patient date shift & mask /\allowbreak{} synthetic replace /\allowbreak{} hash /\allowbreak{} temporal shift \\
    Longitudinal consistency across notes & yes\tblcomma{} cryptographic keyed & not guaranteed across documents \\
    Per-study/\allowbreak{}tenant key isolation & yes & not provided \\
    Distributed execution & Ray GPU/\allowbreak{}CPU\tblcomma{} adaptive batching\tblcomma{} OOM batch splitting & batch processing; no distributed infra \\
    License & MIT & Apache-2.0 \\
    \bottomrule
  \end{tabular}
\end{table}

\begin{table}[H]
  \caption{\textbf{Feature comparison with managed cloud de-identification services.}
  These services require sending clinical text off-site under a business associate
  agreement. AWS Comprehend Medical detects PHI but performs no in-service replacement.
}
  \label{tab:cloud}
  \centering
  \scriptsize
  \resizebox{\textwidth}{!}{%
  \begin{tabular}{P{3.1cm}P{3.1cm}P{3.1cm}P{2.7cm}P{2.5cm}}
    \toprule
    \textbf{Axis} & \textbf{TIDE 2.0} & \textbf{GCP Sensitive Data Protection} & \textbf{Azure Health De-id} & \textbf{AWS Comprehend Medical} \\
    \midrule
    Deployment & on-prem\tblcomma{} open-source (PHI local) & managed cloud & managed cloud & managed cloud \\
    De-identify (replace) or only detect? & detect + replace & detect + redact/\allowbreak{}replace/\allowbreak{}mask/\allowbreak{}crypto/\allowbreak{}tokenize & detect + surrogation & detect only \\
    Longitudinal consistency across notes & yes\tblcomma{} cryptographic keyed & not documented as built in & not documented as built in & n/\allowbreak{}a \\
    Interval-preserving per-patient date shift & yes & generic date-shift\tblcomma{} not per-patient & limited & n/\allowbreak{}a \\
    Model-agnostic recognizer & yes & fixed service models & fixed service models & fixed service models \\
    Batch execution & Ray distributed (single-node benchmark in this paper) & via cloud APIs (egress + quota) & via cloud APIs & via cloud APIs \\
    License /\allowbreak{} openness & MIT\tblcomma{} inspectable & proprietary & proprietary & proprietary \\
    \bottomrule
  \end{tabular}}
\end{table}

\begin{table}[H]
  \caption{\textbf{Decomposition of the SHIELD to i2b2 drop in micro-averaged recall,
  \sentryName{}.} Share: the category's percentage of gold spans on each corpus. Recall
  term: $\bar{w}_c(r_{s,c} - r_{i,c})$, the category's recall change weighted by its mean
  share. Mix term: $\bar{r}_c(w_{s,c} - w_{i,c})$, the effect of the different category
  mix. The two columns sum to the total drop (Methods). A positive value lowers recall
  on i2b2.}
  \label{tab:decomposition}
  \centering
  \small
  \begin{tabular}{lrrrrrr}
    \toprule
    & \multicolumn{2}{c}{\textbf{Share (\%)}} & \multicolumn{2}{c}{\textbf{Recall}} & & \\
    \cmidrule(lr){2-3} \cmidrule(lr){4-5}
    \textbf{Category} & \textbf{SHIELD} & \textbf{i2b2} & \textbf{SHIELD} & \textbf{i2b2} & \textbf{Recall term} & \textbf{Mix term} \\
    \midrule
    AGE & 3.5 & 7.3 & 0.78 & 0.68 & +0.006 & -0.028 \\
    DATE & 34.7 & 45.7 & 0.94 & 0.85 & +0.037 & -0.099 \\
    DOCTOR & 25.4 & 17.6 & 0.90 & 0.85 & +0.010 & +0.069 \\
    HOSPITAL & 9.0 & 8.5 & 0.79 & 0.39 & +0.034 & +0.003 \\
    ID & 7.1 & 6.8 & 0.90 & 0.64 & +0.018 & +0.002 \\
    LOCATION & 4.4 & 4.1 & 0.55 & 0.65 & -0.004 & +0.002 \\
    PATIENT & 10.1 & 8.0 & 0.88 & 0.79 & +0.008 & +0.017 \\
    PHONE & 5.0 & 2.0 & 0.85 & 0.56 & +0.010 & +0.022 \\
    WEB & 0.8 & 0.0 & 0.67 & 0.57 & +0.000 & +0.005 \\
    \textbf{Total} & 100.0 & 100.0 & 0.88 & 0.77 & +0.120 & -0.008 \\
    \bottomrule
  \end{tabular}
\end{table}

\begin{table}[H]
  \caption{\textbf{Effect of the regex layer.} Difference in span-level recall
  ($\Delta$R) and precision ($\Delta$P), \sentryName{}\texttt{+regex} minus
  \sentryName{}, with 95\% intervals from a paired document bootstrap
  (\bootstrapN{} resamples). Micro pools counts over the \shieldNumCats{} categories.}
  \label{tab:regex-gain}
  \centering
  \scriptsize
  \begin{tabular}{lcccc}
    \toprule
    & \multicolumn{2}{c}{\textbf{SHIELD}} & \multicolumn{2}{c}{\textbf{i2b2 2014}} \\
    \cmidrule(lr){2-3} \cmidrule(lr){4-5}
    \textbf{Category} & \textbf{$\Delta$R [95\% CI]} & \textbf{$\Delta$P [95\% CI]} & \textbf{$\Delta$R [95\% CI]} & \textbf{$\Delta$P [95\% CI]} \\
    \midrule
    AGE & +0.000 [+0.000 to +0.000] & +0.000 [+0.000 to +0.000] & +0.000 [+0.000 to +0.000] & +0.000 [+0.000 to +0.000] \\
    DATE & +0.001 [+0.000 to +0.004] & -0.011 [-0.016 to -0.007] & +0.007 [+0.004 to +0.010] & -0.007 [-0.009 to -0.005] \\
    DOCTOR & +0.000 [+0.000 to +0.000] & +0.000 [+0.000 to +0.000] & +0.000 [+0.000 to +0.000] & +0.000 [+0.000 to +0.000] \\
    HOSPITAL & -0.008 [-0.014 to -0.002] & -0.001 [-0.004 to +0.001] & -0.000 [-0.001 to +0.000] & +0.000 [-0.001 to +0.002] \\
    ID & +0.012 [+0.000 to +0.029] & -0.000 [-0.004 to +0.002] & +0.009 [+0.005 to +0.014] & -0.003 [-0.007 to +0.000] \\
    LOCATION & -0.045 [-0.074 to -0.019] & -0.040 [-0.066 to -0.017] & -0.022 [-0.035 to -0.010] & -0.007 [-0.017 to +0.002] \\
    PATIENT & +0.000 [+0.000 to +0.000] & +0.000 [+0.000 to +0.000] & +0.000 [+0.000 to +0.000] & +0.000 [+0.000 to +0.000] \\
    PHONE & +0.000 [+0.000 to +0.000] & +0.000 [+0.000 to +0.000] & -0.002 [-0.006 to +0.000] & +0.003 [-0.001 to +0.009] \\
    WEB & +0.000 [+0.000 to +0.000] & -0.178 [-0.282 to -0.088] & +0.000 [+0.000 to +0.000] & -0.032 [-0.107 to +0.000] \\
    \textbf{Micro} & -0.001 [-0.003 to +0.001] & -0.006 [-0.008 to -0.004] & +0.003 [+0.001 to +0.004] & -0.003 [-0.004 to -0.002] \\
    \bottomrule
  \end{tabular}
\end{table}

\begin{table}[H]
  \caption{\textbf{DOCTOR and PATIENT pooled as NAME.} Span-level precision and recall
  with bootstrap 95\% intervals when the two person categories are scored as one. This
  removes the patient/provider distinction that the external taggers' taxonomies do not
  make.}
  \label{tab:names-merged}
  \centering
  \scriptsize
  \begin{tabular}{lcccc}
    \toprule
    & \multicolumn{2}{c}{\textbf{SHIELD}} & \multicolumn{2}{c}{\textbf{i2b2 2014}} \\
    \cmidrule(lr){2-3} \cmidrule(lr){4-5}
    \textbf{Model} & \textbf{P [95\% CI]} & \textbf{R [95\% CI]} & \textbf{P [95\% CI]} & \textbf{R [95\% CI]} \\
    \midrule
    TIDE2 regex & 0.00 [0.00-0.00] & 0.00 [0.00-0.00] & 0.00 [0.00-0.00] & 0.00 [0.00-0.00] \\
    AIMI v1 & 0.55 [0.53-0.58] & 0.59 [0.56-0.61] & 0.35 [0.34-0.37] & 0.35 [0.33-0.37] \\
    AIMI v2 & 0.53 [0.51-0.55] & 0.56 [0.54-0.58] & 0.33 [0.32-0.35] & 0.33 [0.32-0.35] \\
    TIDE2-Sentry & 0.95 [0.94-0.96] & 0.91 [0.89-0.92] & 0.93 [0.92-0.94] & 0.84 [0.83-0.86] \\
    TIDE2-Sentry + regex & 0.95 [0.94-0.96] & 0.91 [0.89-0.92] & 0.93 [0.92-0.94] & 0.84 [0.83-0.86] \\
    DeBERTa-PII (lakshyakh93) & 0.56 [0.53-0.59] & 0.44 [0.42-0.47] & 0.40 [0.39-0.42] & 0.40 [0.39-0.42] \\
    OpenMed SuperClinical L & 0.81 [0.80-0.83] & 0.73 [0.70-0.75] & 0.81 [0.80-0.82] & 0.76 [0.75-0.78] \\
    OpenMed SuperClinical S & 0.74 [0.72-0.76] & 0.65 [0.62-0.67] & 0.82 [0.80-0.83] & 0.78 [0.77-0.79] \\
    PII-Identifier (antonio) & 0.93 [0.92-0.94] & 0.76 [0.74-0.78] & 0.93 [0.92-0.94] & 0.80 [0.79-0.82] \\
    Ettin-68M Nemotron & 0.94 [0.93-0.95] & 0.81 [0.79-0.84] & 0.93 [0.92-0.94] & 0.82 [0.81-0.83] \\
    ekacare MiniLM & 0.72 [0.69-0.74] & 0.73 [0.71-0.75] & 0.51 [0.50-0.53] & 0.49 [0.47-0.50] \\
    \bottomrule
  \end{tabular}
\end{table}

\end{document}

%% file: tide2_macros.tex
\newcommand{\tidetworepo}{https://github.com/susom/tide2-core}
\newcommand{\sentryName}{TIDE2-Sentry}
\newcommand{\sentryRepo}{https://huggingface.co/stanford-med-hdr/tide2-sentry-clinical-ner}
\newcommand{\sentryBase}{DeBERTa-v3-large}
\newcommand{\sentryTypes}{10}
\newcommand{\teacherName}{Gemini 2.5 Flash}
\newcommand{\teacherLabelNotes}{$\sim$13{,}000}

\newcommand{\windowBudget}{512}
\newcommand{\windowOverlap}{40}
\newcommand{\shieldNotes}{1{,}381}
\newcommand{\shieldGoldSpans}{10{,}229}
\newcommand{\shieldSpansPerNote}{7.4}
\newcommand{\shieldNumCats}{9}
\newcommand{\bootstrapN}{2{,}000}
\newcommand{\overlapThresh}{0.8}

\newcommand{\numExternalPII}{6}
\newcommand{\itwoNotes}{1{,}304}
\newcommand{\itwoGoldSpans}{27{,}298}
\newcommand{\itwoSpansPerNote}{20.9}
\newcommand{\itwoRawSpans}{28{,}867}
\newcommand{\itwoExcludedTags}{PROFESSION and ORGANIZATION}
\newcommand{\itwoAgeExtended}{1{,}390}
\newcommand{\itwoLocMerged}{950}
\newcommand{\thruGpu}{NVIDIA L4}
\newcommand{\thruNumGpu}{1}
\newcommand{\thruVcpus}{16}
\newcommand{\thruRamGb}{64}
\newcommand{\thruPlatform}{GCP VM}

\newcommand{\thruShieldTotalLfour}{1{,}573}

\newcommand{\thruLongTokens}{2{,}096{,}828}
\newcommand{\thruShieldTokens}{359{,}779}

\newcommand{\thruLongVram}{5.8}
\newcommand{\anonStressN}{2{,}000}
\newcommand{\anonDetRate}{100.0\%}

\newcommand{\anonMrnColl}{0.00\%}
\newcommand{\anonNameColl}{0.4\%}
\newcommand{\anonNameCollLarge}{3.3\%}
\newcommand{\anonNameLargeN}{20{,}000}
\newcommand{\anonShortIdCollMin}{35\%}
\newcommand{\anonShortIdCollMax}{37\%}
\newcommand{\anonNamePool}{2{,}276}

%% file: derived_macros.tex
\newcommand{\sentryMicroP}{0.89}
\newcommand{\sentryMicroR}{0.88}
\newcommand{\sentryMacroP}{0.85}
\newcommand{\sentryMacroR}{0.81}

\newcommand{\flashMacroR}{0.90}
\newcommand{\defaultMicroP}{0.88}
\newcommand{\defaultMicroR}{0.88}
\newcommand{\defaultMacroP}{0.83}
\newcommand{\defaultMacroR}{0.80}
\newcommand{\defaultMicroPItwo}{0.87}
\newcommand{\defaultMicroRItwo}{0.77}

\newcommand{\tideOneMicroR}{0.39}
\newcommand{\tideOneMicroP}{0.48}
\newcommand{\sentryMacroRAimiCats}{0.88}
\newcommand{\aimiOneMacroRAimiCats}{0.83}
\newcommand{\sentryMacroRSupItwo}{0.68}
\newcommand{\sentryMacroRAimiCatsItwo}{0.68}
\newcommand{\aimiOneMacroRAimiCatsItwo}{0.73}

\newcommand{\aimiCats}{DATE, DOCTOR, HOSPITAL, ID, PATIENT, PHONE}
\newcommand{\bestOtherName}{AIMI v1}
\newcommand{\bestOtherMicroR}{0.72}
\newcommand{\bestOtherMicroP}{0.67}

\newcommand{\extMicroRMin}{0.30}
\newcommand{\extMicroRMax}{0.51}
\newcommand{\extBestName}{OpenMed SuperClinical S}
\newcommand{\extMicroPBest}{0.50}
\newcommand{\extDoctorRMax}{0.00}
\newcommand{\bestOtherNameItwo}{AIMI v1}
\newcommand{\bestOtherMicroRItwo}{0.67}
\newcommand{\bestOtherMicroPItwo}{0.73}

\newcommand{\bestOtherPatientRItwo}{0.56}

\newcommand{\bestOtherCatsLedItwo}{4}
\newcommand{\extMicroRMinItwo}{0.18}
\newcommand{\extMicroRMaxItwo}{0.48}
\newcommand{\extBestNameItwo}{ekacare MiniLM}
\newcommand{\extMicroPBestItwo}{0.50}
\newcommand{\extDoctorRMaxItwo}{0.00}
\newcommand{\sentryDateR}{0.94}

\newcommand{\sentryDoctorR}{0.90}

\newcommand{\sentryHospR}{0.79}

\newcommand{\sentryIdR}{0.90}

\newcommand{\sentryLocR}{0.55}

\newcommand{\llmFlashLocR}{0.66}

\newcommand{\sentryPhoneR}{0.85}

\newcommand{\sentryMicroPItwo}{0.87}
\newcommand{\sentryMicroRItwo}{0.77}
\newcommand{\sentryMacroPItwo}{0.72}
\newcommand{\sentryMacroRItwo}{0.67}

\newcommand{\sentryDateRItwo}{0.85}

\newcommand{\sentryDoctorRItwo}{0.85}

\newcommand{\sentryHospRItwo}{0.39}

\newcommand{\sentryIdRItwo}{0.64}

\newcommand{\sentryLocRItwo}{0.65}

\newcommand{\sentryPatientRItwo}{0.79}

\newcommand{\sentryPhoneRItwo}{0.56}

\newcommand{\sentryMicroRDropItwo}{0.11}
\newcommand{\sentryHospRDropItwo}{0.39}
\newcommand{\sentryIdRDropItwo}{0.26}
\newcommand{\sentryPhoneRDropItwo}{0.29}

\newcommand{\itwoWebSupportItwo}{7}
\newcommand{\decompDate}{0.037}

\newcommand{\decompHosp}{0.034}
\newcommand{\decompId}{0.018}

\newcommand{\decompPhone}{0.010}

\newcommand{\decompRateTotal}{0.120}
\newcommand{\decompMixTotal}{-0.008}

\newcommand{\decompVocabShare}{56\%}
\newcommand{\decompDateShare}{33\%}
\newcommand{\sentryNameR}{0.91}
\newcommand{\extNameRMin}{0.44}
\newcommand{\extNameRMax}{0.81}
\newcommand{\sentryNameRItwo}{0.84}
\newcommand{\extNameRMinItwo}{0.40}
\newcommand{\extNameRMaxItwo}{0.82}
\newcommand{\regexGainMicroR}{-0.001}
\newcommand{\regexGainMicroRCI}{-0.003 to +0.001}

\newcommand{\regexGainIdR}{+0.012}
\newcommand{\regexGainIdRCI}{+0.000 to +0.029}

\newcommand{\regexGainMicroRItwo}{+0.003}
\newcommand{\regexGainMicroRCIItwo}{+0.001 to +0.004}

\newcommand{\regexGainIdRItwo}{+0.009}
\newcommand{\regexGainIdRCIItwo}{+0.005 to +0.014}

\newcommand{\defaultMicroRIouHalf}{0.90}
\newcommand{\defaultCharR}{0.99}
\newcommand{\defaultMicroRIouHalfItwo}{0.82}
\newcommand{\defaultCharRItwo}{0.95}
\newcommand{\defaultMicroRNoAlignItwo}{0.69}
\newcommand{\regexIdR}{0.56}

\newcommand{\sentryPlusRegexIdR}{0.91}
\newcommand{\distAgeDelta}{-0.12}

\newcommand{\distDateDelta}{-0.04}

\newcommand{\distDoctorDelta}{-0.06}

\newcommand{\distHospDelta}{-0.09}

\newcommand{\distIdDelta}{-0.05}

\newcommand{\distIdPadj}{=0.045}
\newcommand{\distLocDelta}{-0.11}

\newcommand{\distPatientDelta}{-0.07}

\newcommand{\distPhoneDelta}{-0.10}

\newcommand{\distWebDelta}{-0.15}

\newcommand{\distWebPadj}{=0.202}